\documentclass{article}
\usepackage{iclr2027_conference,times}

\usepackage{amsmath,amsfonts,bm}

\def\eqref#1{equation~\ref{#1}}
\def\1{\bm{1}}

\DeclareMathAlphabet{\mathsfit}{\encodingdefault}{\sfdefault}{m}{sl}
\SetMathAlphabet{\mathsfit}{bold}{\encodingdefault}{\sfdefault}{bx}{n}

\usepackage{graphicx}
\usepackage[table,dvipsnames]{xcolor}   % [table] provides \rowcolor (colortbl)
\usepackage{amsmath}
\usepackage{amssymb}
\usepackage{booktabs}
\usepackage{tabularx}
\usepackage{wrapfig}
\usepackage{multirow}
\usepackage{makecell}
\usepackage{adjustbox}
\usepackage{subcaption}                 % subfigure environment (analysis figure)
\usepackage{microtype}
\usepackage{enumitem}
\usepackage{textcomp}                    % \textlangle / \textrangle
\usepackage{tikz}                        % manual divider rules in the overview figures

\usepackage[hidelinks]{hyperref}
\usepackage{url}
\definecolor{midnightgreen}{rgb}{0.0, 0.29, 0.33}

\title{EmbodiedMidtrain: Bridging the Gap between Vision-Language Models
and Vision-Language-Action Models via Mid-training}

\author{%
\textbf{Yiyang Du\textsuperscript{1}, Zhanqiu Guo\textsuperscript{1}, Haochen Zhang\textsuperscript{1}, Annabella Macaluso\textsuperscript{1},}\\
\textbf{Xin Ye\textsuperscript{2}, Liu Ren\textsuperscript{2}, Yonatan Bisk\textsuperscript{1}, Chenyan Xiong\textsuperscript{1}}\\[2pt]
\normalfont\textsuperscript{1}Carnegie Mellon University\\
\normalfont\textsuperscript{2}Bosch Research North America \& Bosch Center for Artificial Intelligence (BCAI)\\[2pt]
\normalfont\texttt{\{yiyangd,cx\}@cs.cmu.edu}%
}

\iclrfinalcopy % Uncomment for the camera-ready version, but NOT for submission.

\begin{document}

\maketitle
\lhead{Preprint. Under review.}

\begin{abstract}
Vision-Language-Action Models (VLAs) inherit their visual and linguistic capabilities from Vision-Language Models (VLMs), yet most VLAs are built from off-the-shelf VLMs that are not adapted to the embodied domain, limiting their downstream performance. In this work, we propose \textit{EmbodiedMidtrain} to bridge the gap between VLMs and VLAs. We first characterize the data distribution gap between them, showing that VLA data occupy compact regions that are largely separated from the broader VLM distribution, while the degree of alignment varies substantially both across and within VLM data sources. Then, we build a mid-training data engine that leverages a lightweight learnable proximity estimator to select the most VLA-aligned candidates from a large VLM pool, and produce \textit{EmbodiedMidtrain-1M}, a VLM dataset that we further use to mid-train the VLM before downstream VLA fine-tuning.
Experiments on both simulated robot manipulation benchmarks and real-world tasks show consistent VLA performance gains over off-the-shelf checkpoints across VLM backbones and VLA action heads. These gains extend to downstream tasks not used to guide data selection.
Further analysis reveals that mid-training provides a stronger initialization for VLA fine-tuning, with gains emerging from the earliest steps and widening throughout training. Moreover, the data engine captures both dataset-level and sample-level alignment signals, favoring spatial reasoning over text-centric tasks while preserving the diversity of the VLM data.
We will release all code, data and models for future research.
\footnote{Project page: \url{https://adu2021.github.io/blog/EmbodiedMidtrain/}}

\end{abstract}

\section{Introduction}
\label{sec:intro}

Recent advances in robotics foundation models have given rise to Vision-Language-Action Models (VLAs), which unify vision perception, language understanding, and action generation within a single model to enable generalist robot control across diverse tasks, environments, and embodiments~\citep{kim2024openvla, black2024pi0, gemini_robotics_2025}. Most VLAs leverage Vision-Language Models (VLMs) as their backbone, aiming to inherit rich visual and linguistic representations that facilitate rapid adaptation to embodied settings~\citep{kim2025oft, pi05, gr00tn1_2025}.

Despite this progress, a fundamental gap exists: most VLAs initialize from general-purpose \emph{off-the-shelf} VLMs that are not tailored toward embodied action generation. VLM pretraining~\citep{llava,pmlr-v202-li23q,Qwen-VL} covers broad vision-language tasks such as captioning, visual question answering, and document understanding, whereas VLA training consists of robotic manipulation trajectories grounded in physical interaction. This distributional mismatch means that, even when a VLM backbone provides a strong initialization for language and visual understanding, its internal representations may not be well-suited for the embodied reasoning that effective action generation demands~\citep{yang2025vlaser}. Bridging this gap remains an open challenge: simply fine-tuning VLMs on curated embodied data does not reliably translate into better VLA performance~\citep{zhang2026vlm4vla}, indicating the need for more principled VLM adaptation to the VLA domain.

\begin{figure}[t]
  \centering
  \newsavebox{\boxa}
  \newsavebox{\boxb}
  \newsavebox{\boxc}
  \newsavebox{\boxd}
  \newlength{\sepwidth}
  \newlength{\figwidtha}
  \newlength{\figwidthb}
  \newlength{\figwidthc}
  \newlength{\figwidthd}
  \newlength{\maximgheight}

  \setlength{\sepwidth}{5pt}
  \setlength{\figwidtha}{0.22\linewidth}
  \setlength{\figwidthb}{0.275\linewidth}
  \setlength{\figwidthc}{0.22\linewidth}
  \setlength{\figwidthd}{\dimexpr\linewidth-\figwidtha-\figwidthb-\figwidthc-3\sepwidth\relax}

  \newcommand{\vsepline}[1]{%
    \begin{tikzpicture}[baseline=(current bounding box.north)]
      \useasboundingbox (0,0) rectangle (\sepwidth,#1);
      \draw[dashed, gray, line width=0.4pt] (0.5\sepwidth,0) -- (0.5\sepwidth,#1);
    \end{tikzpicture}%
  }
  \newcommand{\imgpanel}[2]{%
    \begin{minipage}[t][\maximgheight][c]{#1}
      \centering
      \usebox{#2}
    \end{minipage}%
  }
  \newcommand{\cappanel}[2]{%
    \begin{minipage}[t]{#1}
      \centering
      \small #2
    \end{minipage}%
  }

  \sbox{\boxa}{\includegraphics[pagebox=cropbox, width=\figwidtha]{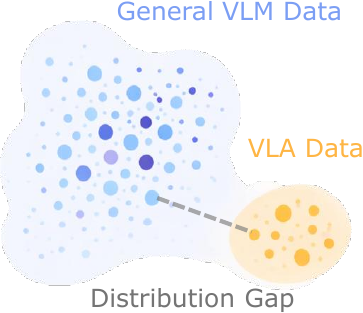}}%
  \sbox{\boxb}{\includegraphics[pagebox=cropbox, width=\figwidthb]{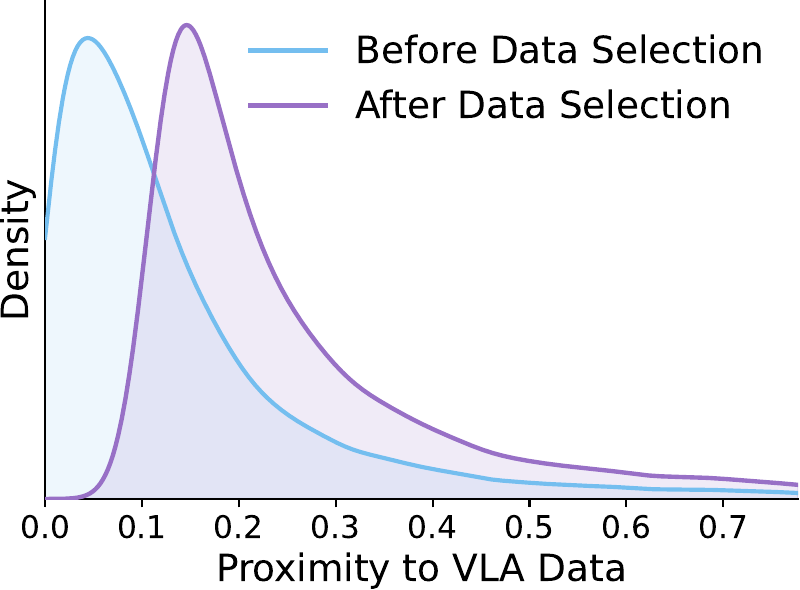}}%
  \sbox{\boxc}{\includegraphics[pagebox=cropbox, width=\figwidthc]{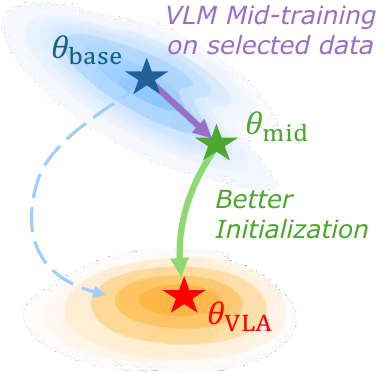}}%
  \sbox{\boxd}{\includegraphics[pagebox=cropbox, width=\figwidthd]{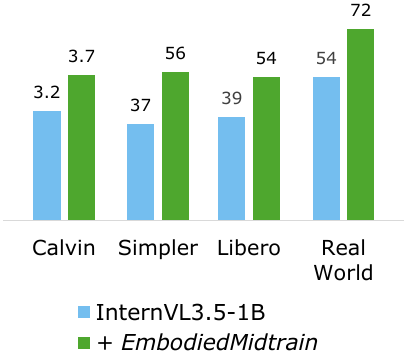}}%

  \setlength{\maximgheight}{\dimexpr\ht\boxa+\dp\boxa\relax}
  \ifdim\dimexpr\ht\boxb+\dp\boxb\relax>\maximgheight
    \setlength{\maximgheight}{\dimexpr\ht\boxb+\dp\boxb\relax}
  \fi
  \ifdim\dimexpr\ht\boxc+\dp\boxc\relax>\maximgheight
    \setlength{\maximgheight}{\dimexpr\ht\boxc+\dp\boxc\relax}
  \fi
  \ifdim\dimexpr\ht\boxd+\dp\boxd\relax>\maximgheight
    \setlength{\maximgheight}{\dimexpr\ht\boxd+\dp\boxd\relax}
  \fi

  \imgpanel{\figwidtha}{\boxa}%
  \vsepline{\maximgheight}%
  \imgpanel{\figwidthb}{\boxb}%
  \vsepline{\maximgheight}%
  \imgpanel{\figwidthc}{\boxc}%
  \vsepline{\maximgheight}%
  \imgpanel{\figwidthd}{\boxd}%

  \par\vspace{0.5em}

  \cappanel{\figwidtha}{(a) VLM-VLA Data Distribution Gap}%
  \hspace{\sepwidth}%
  \cappanel{\figwidthb}{(b) VLM Data\\Distribution Shift}%
  \hspace{\sepwidth}%
  \cappanel{\figwidthc}{(c) VLM Mid-training}%
  \hspace{\sepwidth}%
  \cappanel{\figwidthd}{(d) Downstream\\VLA Gains}%

  \caption{Overview of \textit{EmbodiedMidtrain}.
  We analyze the data distribution gap between VLMs and VLAs, and select VLM samples with higher proximity to the VLA domain for mid-training, yielding a stronger initialization for downstream VLA fine-tuning.}
  \label{fig:overview}
\end{figure}
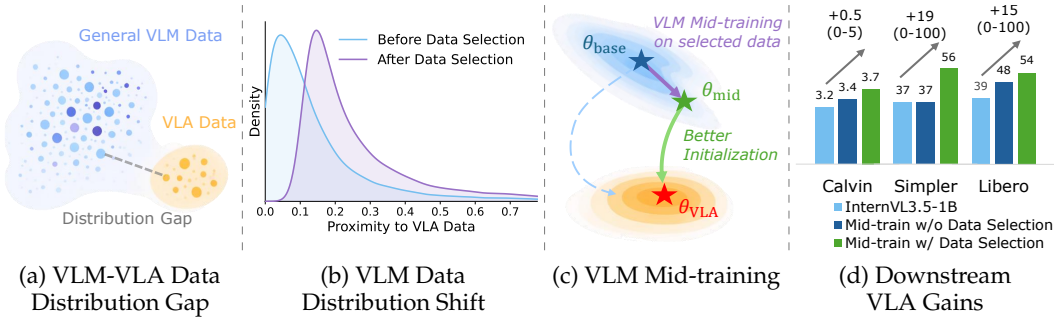

In this work, we present \textit{EmbodiedMidtrain}, a mid-training framework that aligns VLM data distribution with the VLA domain, enabling VLAs for robot manipulation to truly benefit from the abilities of their VLM backbones. We begin with a deep analysis of the VLM-VLA representation gap and identify a core issue~(Figure~\ref{fig:overview}a):
the two domains exhibit a significant data distribution discrepancy. VLA data form compact clusters that are largely separated from the broad, diverse VLM distributions~\citep{xing2025shortcut}; yet this gap is not uniform, as some VLM samples are inherently more aligned with the VLA domain than others. This reveals that the bulk of VLM data concentrates on capabilities distant from VLA data, leaving downstream VLA fine-tuning to bridge a wide representational gap from a misaligned starting point.

Leveraging this understanding, we develop a mid-training data engine that \textbf{reshapes the VLM training distribution toward the VLA domain} before fine-tuning begins~(Figure~\ref{fig:overview}b) to bridge the gap. The key idea is that, rather than training on all available VLM data indiscriminately, we can focus on the subset that is most useful for robotics by aligning with the target VLA distribution, producing a stronger, more relevant initialization for downstream VLA learning. Concretely, we propose a lightweight proximity estimator: a learnable classifier on frozen VLM features that learns to distinguish VLA data from VLM data, and whose predicted scores serve as a continuous measure of each VLM sample's closeness to the VLA domain. Top-scoring samples are then assembled into a curated mid-training mixture, and the VLM is mid-trained on this distribution-aligned dataset to serve as a better initialization for VLA fine-tuning (Figure~\ref{fig:overview}c). This pipeline is lightweight, effective, and requires no architectural changes to the underlying VLM or VLA.

Experiments on Calvin ABC-D~\citep{mees2022calvin}, SimplerEnv Bridge~\citep{walke2023bridgedata}, and Libero-10~\citep{liu2023libero} show that our mid-trained 1.1B model achieves a significant performance boost~(Figure~\ref{fig:overview}d). Beyond raw performance gain, our analysis reveals several insights into what makes mid-training effective.
First, the same curated data transfers across architectures: applying the mixture selected with InternVL3.5-1B to Qwen3VL-2B yields consistent gains, suggesting that proximity-based selection captures transferable properties across VLMs.
Second, the learned proximity estimator substantially outperforms hand-crafted alternatives like feature-space distance and perplexity-based metrics.
Third, training dynamics analysis shows that the mid-trained model outperforms the baseline from the earliest fine-tuning checkpoints and the gap widens over time, indicating mid-training provides a fundamentally better VLM initialization for VLAs.

Our main contributions are as follows:
\begin{itemize}[leftmargin=*,nosep]
\item \textbf{A proximity-based mid-training pipeline and dataset to bridge the VLM-VLA gap.} We propose \textit{EmbodiedMidtrain}, a mid-training pipeline with a data engine that learns a proximity estimator to score VLM samples by their closeness to the VLA domain, and selects the top-ranked candidates to curate a distribution-aligned mid-training dataset \textit{EmbodiedMidtrain-1M}.
\item \textbf{Consistent VLA performance gains across backbones, action heads, and target tasks.} Extensive experiments on simulation benchmarks (Calvin ABC-D, SimplerEnv Bridge, and Libero-10) and real-world tasks demonstrate that VLMs mid-trained on our \textit{EmbodiedMidtrain-1M} dataset yield consistent improvements over the original backbone. We further demonstrate its transferability across VLM backbones, VLA action heads and data selection target tasks.
\item \textbf{Insights on better VLM initialization for VLAs.} We show that mid-training produces an initialization whose advantages emerge from the earliest fine-tuning steps and amplify over time. We further demonstrate that the learned proximity estimation outperforms hand-crafted alternatives by capturing fine-grained alignment with the VLA domain.
\end{itemize}

\section{Related work}

\noindent\textbf{VLA Models.}
VLAs extend foundation models like VLMs to generate robot actions, leveraging the broad visual and language understanding inherited from VLMs~\citep{kim2024openvla,black2024pi0,gr00tn1_2025}.
Existing designs differ in backbone choice and action generation mechanism.
Token-based approaches such as OpenVLA~\citep{kim2024openvla} discretize robot actions into tokens for autoregressive generation; OpenVLA-OFT~\citep{kim2025oft} further introduces parallel decoding with learnable action embeddings for efficient continuous action prediction.
Flow-matching and diffusion-based approaches such as $\pi_0$~\citep{black2024pi0} and $\pi_{0.5}$~\citep{pi05} instead adopt a VLM backbone (PaliGemma~\citep{beyer2024paligemma}) paired with a continuous action generation head.
More recent models such as GR00T N1~\citep{gr00tn1_2025} employ a VLM backbone (Eagle-2~\citep{li2025eagle2}) with a dedicated cross-attention action decoder.
While these designs differ architecturally, a common thread is that the VLM backbone is taken \emph{off-the-shelf} from general-purpose training without specific preparation for the embodied domain, which brings a gap that our work identifies and addresses.

\noindent\textbf{VLM Mid-training.}
Many VLMs adopt multi-stage training pipelines, where the model is further trained on curated vision-language data after initial multimodal pretraining or alignment and before final instruction tuning~\citep{Qwen-VL,Qwen2-VL,internvl25}. In this paper, we refer to this intermediate stage as \emph{VLM mid-training}. Similar to mid-training in LLMs, its goal is to adapt a foundation model toward desired domains or capabilities prior to final post-training or task-specific fine-tuning~\citep{wang2025octothinker,grattafiori2024llama3herdmodels,hu2024minicpmunveilingpotentialsmall,olmo20252olmo2furious}. Our work studies this paradigm in the embodied setting, where mid-training is used to bridge the data distribution gap between general VLM pretraining data and VLA fine-tuning data.

\noindent\textbf{Embodied-oriented VLMs.}
A broad line of work aims to improve VLMs' embodied capabilities, through both embodied-oriented dataset construction~\citep{du-etal-2024-embspatial,zhou2025roborefer,chen2025robo2vlmvisualquestionanswering,pmlr-v270-yuan25c-robopoint} and model-level adaptation~\citep{cai2025spatialbot,ji2025robobrain,internvlam1}.
Despite these efforts, \citet{zhang2026vlm4vla} show that these gains on finetuning VLMs on embodied tasks do not reliably transfer to downstream VLA task performance, suggesting that current embodied VLM fine-tuning captures a different signal from what VLA execution requires.
Alternatively, Vlaser~\citep{yang2025vlaser} converts in-domain robot trajectories into VQA pairs for VLM fine-tuning.
Collectively, these efforts highlight that bridging VLMs and VLAs remains an open problem: existing approaches either improve VLM-side embodied benchmarks without reliably translating into better VLA performance, or require large amounts of in-domain robot data~\citep{yang2025vlaser,RSCL,zhang2026vlm4vla}. Our work takes a complementary perspective by operating on the diverse and abundant VLM data instead of crafting task-specific corpora.
\section{Data distribution gap between VLMs and VLAs}
\label{sec:gap}

\begin{figure}[!t]
  \centering
  \newsavebox{\leftbox}
  \newsavebox{\rightbox}
  \newlength{\maxht}
  \newlength{\maxdp}

  \sbox{\leftbox}{%
    \includegraphics[width=0.53\linewidth]
      {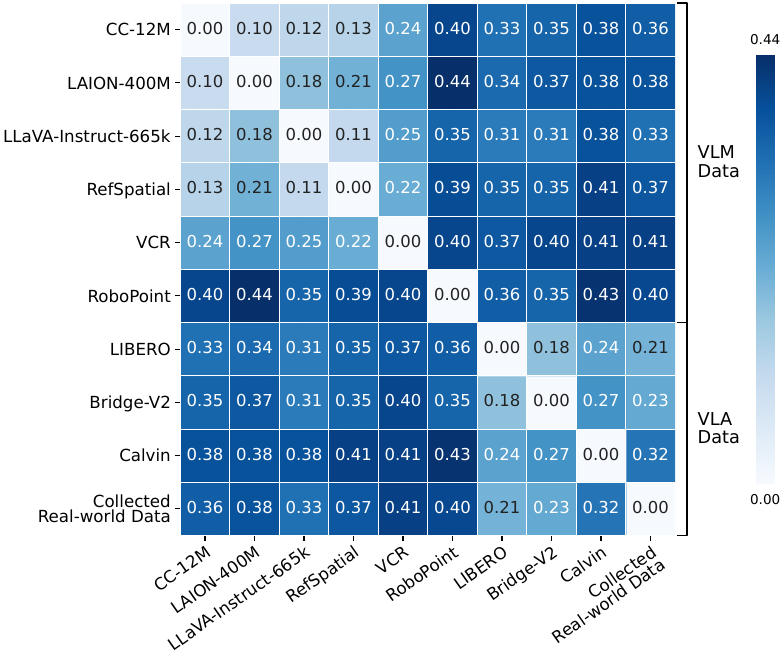}}%
  \sbox{\rightbox}{%
    \includegraphics[width=0.46\linewidth]
      {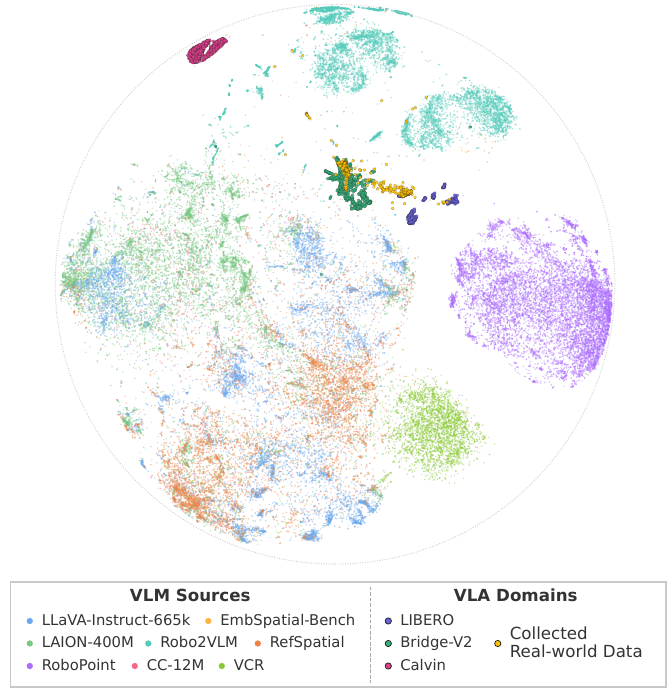}}%

  \pgfmathsetlength{\maxht}{max(\ht\leftbox, \ht\rightbox)}%
  \pgfmathsetlength{\maxdp}{max(\dp\leftbox, \dp\rightbox)}%

  \noindent
  \begin{minipage}[t]{0.535\linewidth}
    \centering
    \makebox[\linewidth][l]{%
  \raisebox{\dimexpr(\maxht-\ht\leftbox)/2\relax}
    [\maxht][\maxdp]{\usebox{\leftbox}}%
}\\[0.5em]
    {\small (a) Pairwise MMD distance matrix between\\
      VLM and VLA datasets}
  \end{minipage}%
  \makebox[0.005\linewidth][c]{%
    \begin{tikzpicture}[baseline={(0,0)}]
      \draw[dashed, line width=0.01pt, gray]
        (0,\maxht) -- (0,-\maxdp);
    \end{tikzpicture}%
  }%
  \begin{minipage}[t]{0.46\linewidth}
    \centering
    \raisebox{\dimexpr(\maxht-\ht\rightbox)/2\relax}
  [\maxht][\maxdp]{\usebox{\rightbox}}\\[0.5em]
    {\small (b) t-SNE visualization of visual feature
      distributions across VLM and VLA datasets}
  \end{minipage}%

  \caption{%
    \textbf{Distribution analysis of VLM and VLA datasets.}
    (a) Pairwise MMD distances quantify this distribution gap,
    with cross-group distances larger than within-group distances.
    (b) VLA datasets (shown with larger markers) form more compact and concentrated clusters
    that are separated from the broader and more dispersed
    VLM distributions.
  }
  \label{fig:tsne-mmd}
  \vspace{-1.5em}
\end{figure}
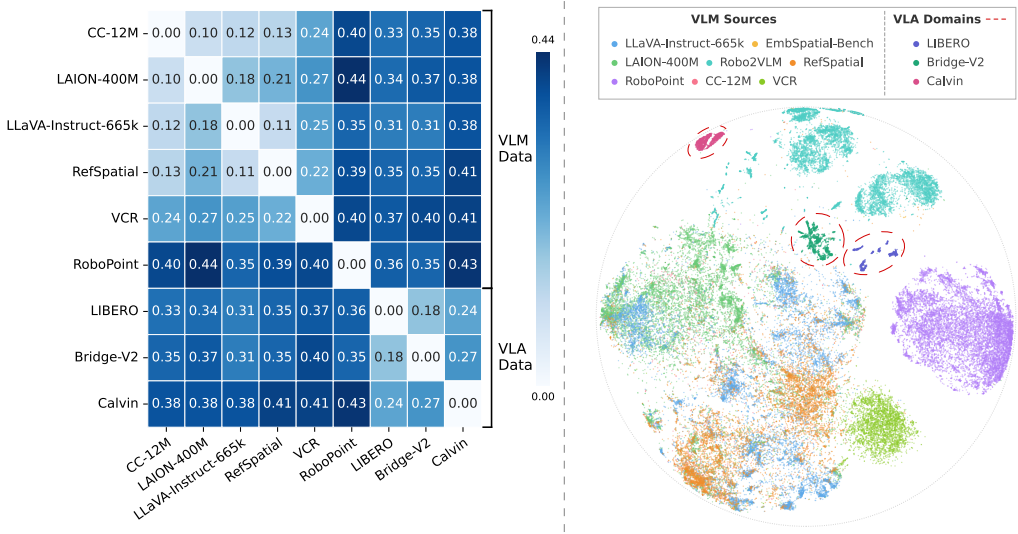

Since most VLAs inherit their visual and linguistic representations from VLM pretraining, the quality of this initialization is fundamentally shaped by the data the VLM was trained on. This motivates us to examine the gap between VLMs and VLAs from the perspective of their training data distributions, where we analyze VLM and VLA data in a shared representation space and characterize the gap with both qualitative and quantitative measurements.

To represent VLM and VLA data in a unified distribution space, we extract the last hidden states of InternVL3.5 VLM~\citep{wang2025internvl3_5} as the feature representation $h(\cdot)$ for each data sample. We first quantify the distance between each pair of datasets with \textbf{Maximum Mean Discrepancy} (MMD).
Formally, the squared MMD between two datasets $P$ and $Q$ is:
\begin{equation*}
\begin{aligned}
\operatorname{MMD}^2(P,Q)
={}& \mathbb{E}_{x,x' \sim P}[k(x,x')] + \mathbb{E}_{y,y' \sim Q}[k(y,y')] 
   - 2\,\mathbb{E}_{x \sim P,\, y \sim Q}[k(x,y)]
\end{aligned}
\end{equation*}
where $k(x,y) = \exp\!\bigl(-\lVert h(x) - h(y) \rVert_2^2 \,/\, 2\sigma^2\bigr)$ is a Gaussian RBF kernel applied over the extracted features, with bandwidth $\sigma$ set via the median heuristic~\citep{gretton2012kernel}. Figure~\ref{fig:tsne-mmd}a reports the globally normalized pairwise MMD scores across all dataset pairs. We further visualize these features using t-SNE~\citep{JMLR:v9:vandermaaten08a} in Figure~\ref{fig:tsne-mmd}b.

\noindent\textbf{The characteristics of data distribution gap between VLM and VLA.} The distribution space of VLM data and VLA data is largely separated, with only a few near-neighbors. Figure~\ref{fig:tsne-mmd}a shows that MMD distances are generally smaller within the VLM group and within the VLA group than across the two groups, quantitatively confirming a clear distributional mismatch. Figure~\ref{fig:tsne-mmd}b further illustrates this pattern: VLA datasets form compact clusters that are mostly detached from the main regions occupied by VLM datasets, while only a small subset of the VLM data lies nearby. These results reveal a clear distribution mismatch between VLM data and VLA data that could affect the performance of downstream embodied tasks.

\noindent\textbf{The data distribution gap exhibits substantial internal heterogeneity.}  Although VLM and VLA data are separated overall, their mismatch is not uniform within each dataset. Figure~\ref{fig:tsne-mmd}b shows that part of VLM sources lie noticeably closer to VLA domains than others, with a few local regions exhibiting clear cross-domain proximity despite the global separation. This suggests that the gap between VLM and VLA is better characterized as a spectrum of alignment, revealing substantial heterogeneity in how different parts of the VLM distribution relate to VLA data.

These insights suggest that \textbf{enhancing VLMs for embodied tasks requires reshaping the training data distribution toward the VLA domain}. Realizing this goal requires more than coarse dataset-level mixture adjustment; it calls for sample-wise selection within each dataset.
This motivates a data-centric mid-training strategy that explicitly prioritizes VLM samples that are most compatible with the target embodied domain.

\section{Data Engine for \textit{EmbodiedMidtrain}}
\label{sec:midtrain}

The distribution analysis in Section~3 reveals substantial differences
between VLM and VLA data, together with heterogeneity across and within
VLM sources. These observations motivate a mid-training data engine
that prioritizes VLM examples according to their affinity to the target
VLA domain. We construct the candidate pool from both general-purpose
and embodied-oriented VLM sources to provide varied training examples,
and perform selection at the sample level to account for variation
within each source.

\noindent\textbf{VLM data source.}
To support mid-training that bridges the gap between VLM capabilities and downstream VLA performance, we first construct a candidate pool spanning both general-purpose and embodied-oriented VLM datasets.
For \textbf{general VLM data}, following standard VLM mid-training data of Qwen-VL~\citep{Qwen-VL}, we include image-captioning data from subsets of LAION-400M~\citep{laion400m}, CC-12M~\citep{changpinyo2021cc12m} with captions relabeled by BLIP~\citep{li2022blip}, instruction-following data from LLaVA-Instruct-665k~\citep{liu2024improvedbaselinesvisualinstruction}, and Visual Commonsense Reasoning (VCR)~\citep{zellers2019vcr}. For \textbf{embodied-oriented VLM data}, we additionally include RefSpatial~\citep{zhou2025roborefer} for spatial referring and reasoning, evaluation examples from EmbSpatial-Bench~\citep{du-etal-2024-embspatial} for embodied spatial understanding, Robo2VLM~\citep{chen2025robo2vlmvisualquestionanswering} for robotic visual question answering, and RoboPoint~\citep{pmlr-v270-yuan25c-robopoint} for spatial affordance prediction.

\paragraph{Proximity-Based Data Selection.}
Inspired by target-guided data selection for LLM pretraining
\citep{xie2023data}, we frame sample scoring as a domain-membership
problem. A lightweight binary classifier trained on frozen VLM features
distinguishes target VLA samples from candidate VLM samples. We use its
predicted probability of VLA membership as a proximity score for
ranking candidate examples.

Let $D_{\mathrm{VLM}}$ and $D_{\mathrm{VLA}}$ denote the candidate VLM
pool and target VLA training corpus, respectively. Given a selection
budget $K$ and a scoring function $s$, we select the subset with the
highest aggregate proximity score:
\begin{equation}
    D_{\mathrm{sel}}
    =
    \underset{
        \substack{
            D' \subseteq D_{\mathrm{VLM}} \\
            |D'| = K
        }
    }{\arg\max}
    \sum_{x \in D'} s(x).
    \label{eq:midtrain_selection_objective}
\end{equation}
Because the objective is additive across samples, its solution is
obtained by retaining the $K$ highest-scoring candidates:
\begin{equation}
    D_{\mathrm{sel}}
    =
    \operatorname{TopK}_{x \in D_{\mathrm{VLM}}} s(x).
    \label{eq:midtrain_topk}
\end{equation}

The key question is how to define $s$. Let $\phi(x)$ denote the frozen
VLM representation of sample $x$, and let $p_{\mathrm{VLM}}$ and
$p_{\mathrm{VLA}}$ denote the two domain densities in this representation
space. The density ratio
$p_{\mathrm{VLA}}(\phi(x)) / p_{\mathrm{VLM}}(\phi(x))$
measures the relative likelihood of a representation under the target
domain compared with the candidate distribution.

A classical result connects this ratio to binary classification
\citep{GAN}. Under equal domain priors, minimizing
the population binary cross-entropy objective over unrestricted scoring
functions yields the optimal domain-membership score:
\begin{equation}
    s^{*}(x)
    =
    \frac{
        p_{\mathrm{VLA}}(\phi(x))
    }{
        p_{\mathrm{VLA}}(\phi(x))
        +
        p_{\mathrm{VLM}}(\phi(x))
    }.
    \label{eq:midtrain_optimal_score}
\end{equation}
This score is monotonically increasing in the density ratio. The result
motivates using a learned classifier as a practical proxy for relative
domain likelihood; the downstream usefulness of the resulting selection
is evaluated empirically.

We instantiate the proximity estimator as a linear scoring function $f$
on the frozen VLM's last-token, last-hidden-state representation,
followed by a sigmoid:
\begin{equation}
    s(x) = \sigma\!\left(f(\phi(x))\right).
    \label{eq:midtrain_learned_score}
\end{equation}
We train the estimator using VLA training samples as positives and
candidate VLM samples as negatives, with equal weighting of the two
domains:
\begin{equation}
    \mathcal{L}_{\mathrm{cls}}
    =
    -\mathbb{E}_{y \sim D_{\mathrm{VLA}}}
        \left[\log s(y)\right]
    -\mathbb{E}_{x \sim D_{\mathrm{VLM}}}
        \left[\log\!\left(1-s(x)\right)\right].
    \label{eq:midtrain_classifier_loss}
\end{equation}

After training, we score all candidate VLM samples and retain the top
$K$ according to Equation~\ref{eq:midtrain_topk}. This produces
\textit{EmbodiedMidtrain-1M}, the corpus used for subsequent VLM
mid-training. The selection procedure prioritizes high-scoring examples
from the heterogeneous candidate pool, with its downstream effectiveness
and selected-data characteristics examined in Section~\ref{sec:analysis}.
\section{Experiments}

We conduct experiments to demonstrate that the proposed mid-training data engine bridges the VLM-VLA gap. In particular, we show that mid-training with proximity-based data selection consistently improves multiple VLM backbones on downstream VLA tasks, and that the curated data mixture exhibits cross-backbone transferability.

\subsection{Setup}
\label{sec:setup}

\noindent\textbf{VLM mid-training.}
We first construct the mid-training dataset using the data engine described in Section~\ref{sec:midtrain}.
We leverage a balanced mixture of training data in VLA fine-tuning as the target VLA data for proximity estimator learning.
To prevent overfitting, we apply early stopping at 90\% validation accuracy. The resulting proximity estimator scores and selects candidate VLM samples to conduct mid-training on InternVL3.5-1B~\citep{wang2025internvl3_5} and Qwen3VL-2B~\citep{Qwen3-VL} using the LLaMA-Factory~\citep{zheng2024llamafactory} framework.
We train all model parameters with a global batch size of 256 for 5,000 steps. See Appendix~\ref{app:impl_detail_vlm} for more implementation details.

\noindent\textbf{VLA fine-tuning.}
After VLM mid-training, we initialize a VLA from the resulting VLMs and fine-tune it following the VLA training pipeline of VLM4VLA~\citep{zhang2026vlm4vla}, which enables controlled comparisons of VLM initializations by fixing the downstream robot data and using a shared lightweight two-branch MLP action head.
By contrast, models such as $\pi_{0.5}$~\citep{pi05} introduce large-scale robot pretraining, making the contribution of VLM initialization harder to isolate.
The architecture and training details are provided in Appendix~\ref{app:impl_detail_vla}.

\noindent\textbf{Evaluation benchmarks.} We evaluate on three simulated manipulation benchmarks that cover multiple aspects of embodied control, following VLM4VLA's evaluation protocol for fair and reproducible comparison. \textbf{Calvin ABC-D}~\citep{mees2022calvin} trains on the ABC splits and evaluates on the unseen split D over 1,000 five-subtask sequences, testing generalization to novel scene configurations. \textbf{SimplerEnv Bridge}~\citep{li2025simpler,walke2023bridgedata} is a real-to-sim benchmark with four tabletop manipulation tasks and 24 randomized trials each, reporting mean success rate. \textbf{LIBERO-10}~\citep{liu2023libero}, the most challenging suite in the LIBERO benchmark, comprises 10 long-horizon tasks evaluated on 50 trials per task.

\subsection{Baselines}

We compare our method against two categories of baselines: expert VLA models and VLAs trained from off-the-shelf VLMs. These baselines are experimented under the framework of VLM4VLA for a unified comparison.
Additional implementation details are provided in the Appendix~\ref{app:impl_detail_vla}.

\noindent\textbf{Expert VLA Baselines.}
We first compare with representative expert VLA models, including OpenVLA~\citep{kim2024openvla} and $\pi_0$~\citep{black2024pi0}. OpenVLA is built on Llama-2-7B with DINOv2 and SigLIP visual encoders, and models robot control by autoregressively predicting discretized action tokens. $\pi_0$ is based on Paligemma-1 VLM and models continuous robot actions using flow matching. We adapt VLM4VLA's implementation on the architecture of OpenVLA and $\pi_0$ and apply the same training recipe to make results comparable.

\noindent\textbf{Off-the-shelf VLM Baselines.}
To evaluate how pretrained VLM abilities transfer to downstream robot manipulation tasks, we further consider VLAs directly finetuned from a diverse set of off-the-shelf VLMs spanning different architectures and scales. These VLMs include the Qwen2.5VL and Qwen3VL family~\citep{Qwen2.5-VL,Qwen3-VL}, which represents a strong line of general-purpose open-source VLMs; the Paligemma family (Paligemma-1~\citep{beyer2024paligemma} and Paligemma-2~\citep{steiner2024paligemma2}), which is widely adopted for downstream adaptation; and KosMos-2~\citep{peng2023kosmos2}, which provides a grounding-oriented alternative.

\begin{table}[tbp]
\centering
\caption{Main results across Calvin ABC-D, SimplerEnv-Bridge, and Libero-10. \textit{EmbodiedMidtrain} consistently improves the performance of downstream VLA benchmarks for both Qwen3-VL-2B and InternVL3.5-1B backbones. * Results for Expert VLA Baselines and Off-the-shelf VLM Baselines are reported by VLM4VLA under the same framework.}
\label{tab:main_results}
\small
% Match the table height before removing the training-budget column.
\renewcommand{\arraystretch}{0.88}
\setlength{\tabcolsep}{2pt}
\begin{tabular*}{\linewidth}{@{\extracolsep{\fill}}l r c c c c c c c c@{}}
\toprule
\multirow{2}{*}{\textbf{Model}} & \multirow{2}{*}{\textbf{Size}} & \multicolumn{6}{c}{\textbf{Calvin (Tasks Completed in a Row)}} & \multirow{2}{*}{\textbf{Simpler$\uparrow$}} & \multirow{2}{*}{\textbf{Libero$\uparrow$}} \\
& & 1 & 2 & 3 & 4 & 5 & \makecell{\textbf{Avg.}\\\textbf{Len.$\uparrow$}} & & \\
\midrule
\rowcolor{gray!20}\multicolumn{10}{c}{\textbf{Expert VLA Baselines*}}\\
\midrule
OpenVLA (Llama-2) & 7.7B & 0.792 & 0.644 & 0.499 & 0.368 & 0.245 & 2.548 & \phantom{0}4.2 & 53.7 \\
$\pi_0$ (Paligemma-1) & 3.1B & 0.896 & 0.785 & 0.686 & 0.610 & 0.532 & 3.509 & 60.4 & 46.0 \\
\midrule
\rowcolor{gray!20}\multicolumn{10}{c}{\textbf{Off-the-shelf VLM Baselines*}}\\
\midrule
Qwen2.5VL-3B & 3.8B & 0.922 & 0.842 & 0.766 & 0.700 & 0.626 & 3.856 & 48.0 & 43.0 \\
Qwen2.5VL-7B & 8.3B & 0.935 & 0.864 & 0.807 & 0.758 & 0.693 & 4.057 & 46.8 & 45.0 \\
\midrule
Qwen3VL-4B & 4.4B & 0.933 & 0.857 & 0.790 & 0.719 & 0.644 & 3.943 & 56.3 & 44.4 \\
Qwen3VL-8B & 8.8B & 0.940 & 0.868 & 0.797 & 0.746 & 0.684 & 4.035 & 58.3 & 46.2 \\
Qwen3VL-30B-A3B & 30B-A3B & 0.939 & 0.877 & 0.820 & 0.757 & 0.682 & 4.075 & 44.8 & 46.8 \\
\midrule
Paligemma-1 & 2.9B & 0.914 & 0.813 & 0.692 & 0.599 & 0.488 & 3.506 & 55.3 & 44.2 \\
Paligemma-2 & 3.0B & 0.901 & 0.775 & 0.669 & 0.575 & 0.486 & 3.406 & 57.3 & 46.2 \\
KosMos-2 & 1.7B & 0.878 & 0.721 & 0.591 & 0.498 & 0.408 & 3.096 & 60.4 & 55.0 \\
\midrule
\rowcolor{gray!20}\multicolumn{10}{c}{\textbf{VLM with \textit{EmbodiedMidtrain} (Ours)}}\\
\midrule
Qwen3VL-2B & 2.1B & 0.887 & 0.747 & 0.612 & 0.527 & 0.432 & 3.205 & 38.5 & 33.8 \\
\quad \textit{+ EmbodiedMidtrain} & 2.1B & 0.922 & 0.808 & 0.700 & 0.623 & 0.533 & 3.584 & 45.8 & 40.2 \\
InternVL3.5-1B & 1.1B & 0.909 & 0.754 & 0.606 & 0.498 & 0.406 & 3.173 & 36.5 & 39.0 \\
\quad \textit{+ EmbodiedMidtrain} & 1.1B & 0.935 & 0.838 & 0.737 & 0.653 & 0.551 & \textbf{3.714} & \textbf{56.3} & \textbf{54.2} \\
\bottomrule
\end{tabular*}

\end{table}

\subsection{Results}
\label{sec:results}

Table~\ref{tab:main_results} presents the main results. With proximity-based mid-training, our models achieve consistent and substantial improvements across all three benchmarks, while finetuned on the same robot action data, demonstrating that a well-curated mid-training mixture can significantly strengthen a VLM's readiness for downstream action learning.
More details are provided in Appendix~\ref{app:detailed_results}.

\noindent\textbf{Consistent performance gains across benchmarks.}
\textit{EmbodiedMidtrain} improves both InternVL3.5-1B and Qwen3VL-2B on all three benchmarks relative to their off-the-shelf initializations. For InternVL3.5-1B, mid-training increases the average number of consecutively completed tasks on Calvin ABC-D by $0.541$ and the success rates on SimplerEnv Bridge and LIBERO-10 by $19.8$ and $15.2$ percentage points, respectively.

\noindent\textbf{Transferability across VLM backbones.}
To assess transfer across backbones, we mid-train Qwen3VL-2B on the same data selected using InternVL3.5-1B's feature space. The gains on Qwen3VL-2B are therefore obtained without training a new proximity estimator or reselecting data for this backbone. This suggests that proximity-based selection captures aspects of alignment with downstream VLA tasks that transfer across VLM architectures.

\subsection{Real-World Experiments}
\label{sec:real_world_exp}

\begin{wrapfigure}{r}{0.5\textwidth}
\vspace{-\intextsep}
\centering
\includegraphics[width=\linewidth]{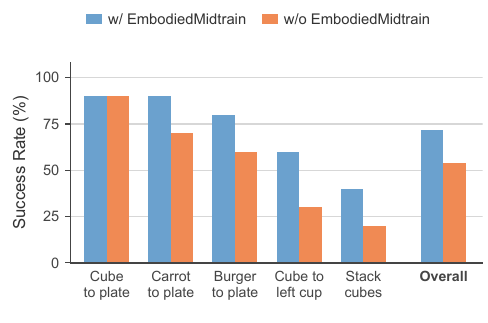}
\caption{Real-world success rates with and without \textit{EmbodiedMidtrain} across five manipulation tasks on InternVL-3.5-1B backbone.}

\label{fig:realworld_results}
\vspace{-2em}
\end{wrapfigure}

We further demonstrate that the gains from mid-training transfer to real-world manipulation tasks. We collect 50 demonstrations for each of five manipulation tasks and evaluate with 10 rollouts per task on a Franka Research 3 robot with a Robotiq 2F-85 gripper. As shown in Figure~\ref{fig:realworld_results}, \textit{EmbodiedMidtrain} increases the overall success rate from 54\% to 72\% across five tasks, improving performance on four real-world tasks.

For real-world experiments, we initialize the VLA with the same mid-trained VLM checkpoint used for the simulation experiments in Section~\ref{sec:results}, then fine-tune it on the collected real-world demonstrations. These results show that \textbf{\textit{EmbodiedMidtrain} can select mid-training data using simulation data and still improve performance on different real-world manipulation tasks}. Details and qualitative examples are provided in Appendices~\ref{app:impl_detail_real_world} and~\ref{app:qualitative}.

Additional experiments in Appendix~\ref{app:transferability} show transferability across VLA action heads and target tasks: downstream tasks not used to guide data selection still benefit from \textit{EmbodiedMidtrain}. Together with the gains across VLM backbones, these findings suggest that the mid-training provides better initialization broadly transferrable on embodied manipulation tasks.
\section{Analysis}
\label{sec:analysis}

Having established the effectiveness of our approach, we now examine its design choices and behavior in detail. We first ablate two core components of our data engine: the effect of data selection versus random sampling, and the choice of proximity measurement (Section~\ref{sec:ablations}). We then analyze the training dynamics of VLA fine-tuning to understand \emph{when} the benefits of mid-training emerge (Section~\ref{sec:training_dynamics}), and finally inspect the selected data to understand \emph{what} the proximity estimator learns to prefer (Section~\ref{sec:analysis_selected}).

\subsection{Ablations}
\label{sec:ablations}

\begin{wraptable}{r}{0.51\textwidth}
\vspace{-1em}
\centering
\caption{Ablation results for random selection and different proximity measurements on mid-training InternVL3.5-1B backbone.}
\label{tab:ablations}
\footnotesize
\setlength{\tabcolsep}{2.5pt}
\begin{tabularx}{\linewidth}{@{}>{\raggedright\arraybackslash}X c c c@{}}
\toprule
\textbf{Setting} & \textbf{Calvin$\uparrow$} & \textbf{Simpler$\uparrow$} & \textbf{Libero$\uparrow$} \\
\midrule
Random Selection & 3.398 & 43.8 & 48.4 \\
\midrule
\rowcolor{gray!20}\multicolumn{4}{l}
{\textit{Proximity Measurements}} \\
\midrule
Feat.-space Avg. Dist. & 3.126 & 53.1 & 51.2 \\
VLA-cond. Perplexity & 3.159 & 55.2 & 48.0 \\
Delta Perplexity & 3.221 & 39.6 & 54.2 \\
Learned Estimator (\textit{Ours}) & \textbf{3.714} & \textbf{56.3} & \textbf{54.2} \\
\bottomrule
% \vspace{-em}
\end{tabularx}

\end{wraptable}

We ablate two central design choices in our data engine: the advantage of proximity-based selection over random sampling, and the effectiveness of different proximity measurements. Table~\ref{tab:ablations} summarizes the results.

\noindent\textbf{Random selection.}
Randomly sampling from the full candidate pool consistently underperforms our learned estimator on all three benchmarks, showing that naive mid-training on unfiltered data is insufficient to bridge the distribution gap. This suggests that the gains do not come merely from additional mid-training, but from identifying and retaining the subset of VLM data that is better aligned with the VLA domain. Proximity-based selection is therefore critical to unlocking the benefit of mid-training.

\noindent\textbf{Proximity measurements.}
Beyond our learned proximity estimator, we evaluate three alternatives: \emph{feature-space average distance} (mean distance to VLA samples in the frozen VLM's representation space), \emph{VLA-conditioned perplexity} (perplexity under a VLM fine-tuned on text-form VLA data), and \emph{delta perplexity} (perplexity change relative to the original VLM). All alternatives are less consistent than our learned estimator, which confirms that hand-crafted metrics only partially capture VLA alignment, whereas the learned estimator provides a more robust and transferable signal. Formal definitions and details of each alternative are provided in Appendix~\ref{app:impl_proximity}.

\subsection{Training dynamics}
\label{sec:training_dynamics}

\begin{figure}[!t]
    \centering
    \includegraphics[width=\linewidth]{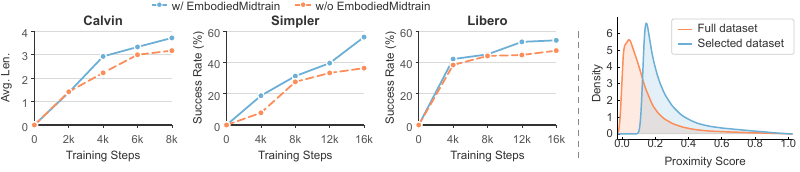}
    \caption{The training dynamics across VLA tasks for VLMs with and without \textit{EmbodiedMidtrain}, showing downstream VLA task performances (left three panels) alongside the VLM data distribution shift of proximity score before and after data selection (right).}
    \label{fig:training_dynamics}
    \label{fig:score_distribution_overall}
\end{figure}

Figure~\ref{fig:training_dynamics} compares VLA fine-tuning trajectories from the original InternVL3.5-1B backbone and our mid-trained variant, evaluated at intermediate checkpoints throughout training. The mid-trained model already achieves higher performance in the early stages of fine-tuning, providing direct evidence that \textbf{proximity-based mid-training yields a better initialization for VLA learning}. Moreover, this advantage is not a transient head start, since the mid-trained model consistently outperforms the baseline throughout the entire training trajectory, and the gap widens rather than narrowing over time.
Together with the consistently stronger downstream performance of the mid-trained model, these results indicate that the benefits of mid-training persist throughout VLA fine-tuning and translate into a backbone better suited to VLA tasks.

\subsection{Analysis of selected VLM data}
\label{sec:analysis_selected}

\newlength{\analysispanelht}
\setlength{\analysispanelht}{5.5cm}

\begin{figure}[t]
\centering
    \begin{subfigure}[t]{0.4\linewidth}
        \centering
        \begin{minipage}[c][\analysispanelht][c]{\linewidth}
            \centering
            \includegraphics[width=\linewidth]{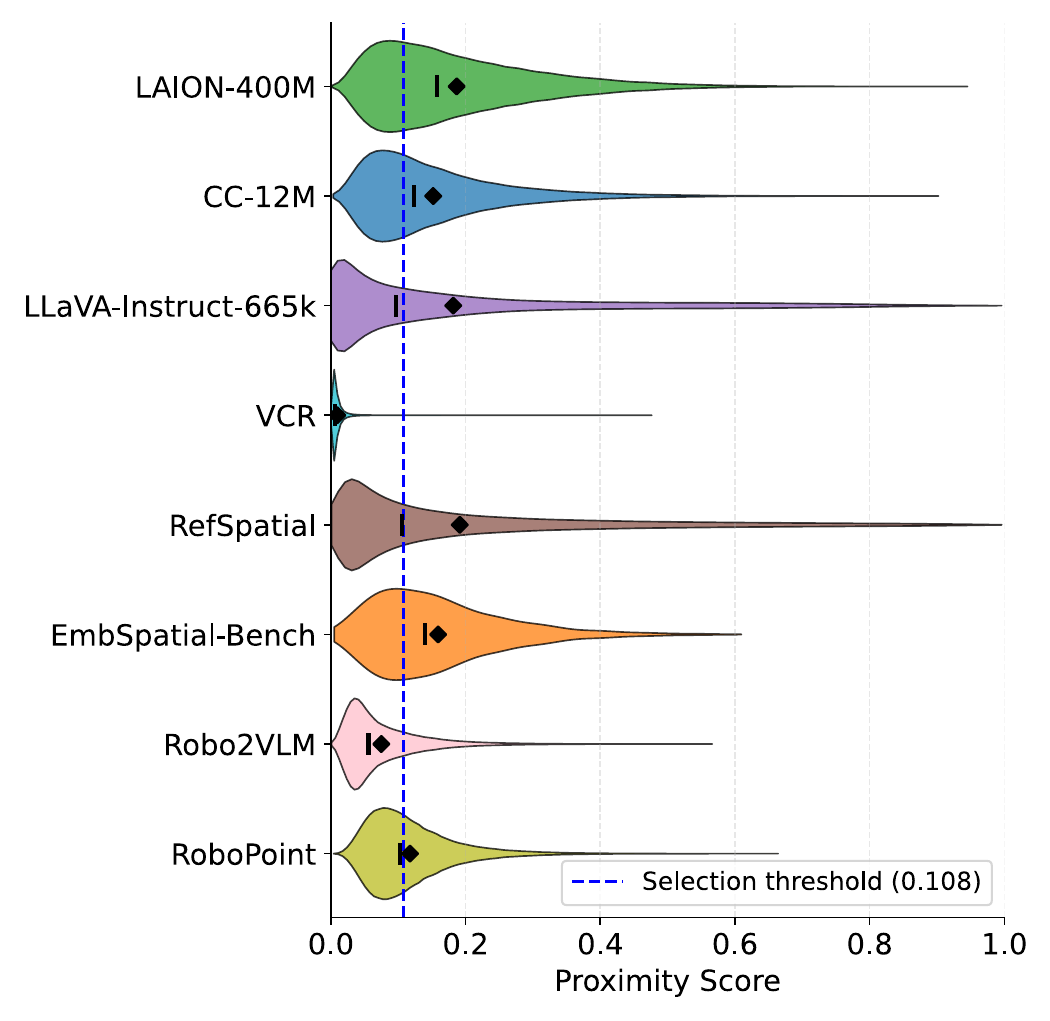}
        \end{minipage}
        \caption{Per-dataset proximity score}
        \label{fig:score_distribution}
    \end{subfigure}
    \hfill
    \begin{subfigure}[t]{0.28\linewidth}
        \centering
        \begin{minipage}[c][\analysispanelht][c]{\linewidth}
            \centering
            \fbox{\includegraphics[width=0.93\linewidth]{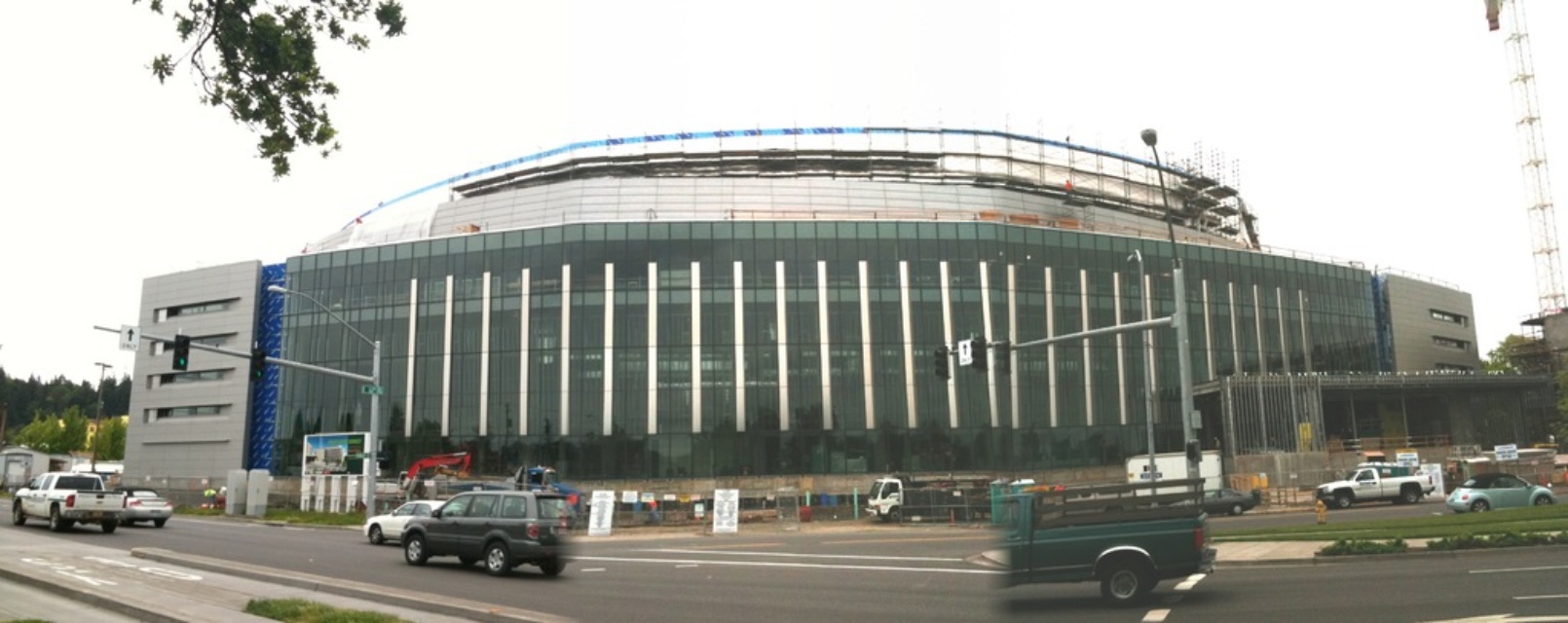}}\par
            \vspace{1ex}
            \scriptsize\raggedright
            
            \textbf{Q:} \textit{You are standing at the point marked by the coordinate point at point (0.878, 0.780). Which object is directly in front of you?}\\
            \textbf{A:} \textit{The white matte truck at lower right.}\\[1ex]
            
            \textbf{Q:} \textit{Locate a point on the yellow metallic crane at upper right. \textnormal{\textlangle format instructions...\textrangle}}\\
            \textbf{A:} \texttt{[(0.976, 0.244)]}
        \end{minipage}
        \caption{High-scoring sample}
        \label{fig:case_positive}
    \end{subfigure}
    \hfill
    \begin{subfigure}[t]{0.28\linewidth}
        \centering
        \begin{minipage}[c][\analysispanelht][c]{\linewidth}
            \centering
            \fbox{\includegraphics[height=2.6cm]{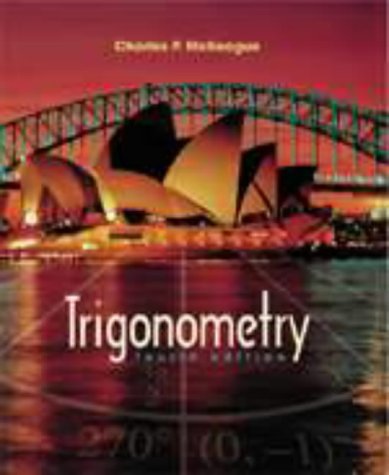}}\par
            \vspace{1ex}
            \scriptsize\raggedright
            \textbf{Q:} \textit{Who wrote this book?}\\
            \textbf{A:} \textit{Charles P. McKeague.}\\[1ex]
            \textbf{Q:} \textit{What is the title of this book?}\\
            \textbf{A:} \textit{Trigonometry.}
        \end{minipage}
        \caption{Low-scoring sample}
        \label{fig:case_negative}
    \end{subfigure}
\caption{Analysis of proximity-based data selection. (a)~Distribution of proximity scores across VLM data sources. (b)~A high-scoring sample from RefSpatial requiring spatial grounding and reasoning. (c)~A low-scoring sample: a book cover with text-only VQA.}
\label{fig:analysis}
\end{figure}

Figure~\ref{fig:score_distribution} presents the proximity score distributions across eight candidate VLM datasets, visualized as violin plots. While all datasets concentrate in the low-to-moderate score range, the distribution shapes vary noticeably across datasets. Among them, RefSpatial achieves the highest average scores while VCR receives the lowest, indicating that the estimator assigns clear \emph{dataset-level} preferences. At the same time, the within-dataset score spread shows that the estimator also performs fine-grained \emph{sample-level} selection, retaining only the most VLA-aligned samples even from high-scoring datasets.

Figures~\ref{fig:case_positive} and~\ref{fig:case_negative} illustrate this with representative examples. The high-scoring sample requires grounding spatial references and exercising spatial reasoning crucial to embodied manipulation. The low-scoring sample involves recognizing text on a book cover, bearing little relevance to embodied tasks. This suggests that the proximity estimator learns to distinguish visual reasoning patterns that transfer to embodied control from those that do not, without requiring manual domain expertise.

The overall proximity score distribution before and after data selection is presented in the rightmost panel of Figure~\ref{fig:score_distribution_overall}. Compared with the full candidate pool, the selected subset exhibits a distribution shift toward higher proximity scores, with part of the low-score mass removed, showing that the reshaped data distribution is more aligned with the VLA domain.
Further analyses of dataset composition and diversity preservation are provided in Appendix~\ref{app:selected_mixture}.

\section{Conclusion}

We presented \textit{EmbodiedMidtrain}, a mid-training pipeline that bridges the data distribution gap between VLMs and VLAs through proximity-based data selection. Built on frozen VLM features, our lightweight estimator identifies the VLM samples most aligned with the VLA domain and uses them to construct a more effective mid-training set. Across simulation benchmarks and real-world tasks, this strategy consistently improves performance competitive with models trained at larger scales and training budgets. The gain transfers across VLM backbones, VLA action heads and target tasks.

Our analysis further shows that mid-training provides a stronger initialization for downstream VLA learning: the advantage emerges from the earliest stages of fine-tuning and continues to grow throughout training. We also find that the selected data exhibit meaningful preferences at both the dataset and sample levels, while preserving the diversity of the original VLM dataset. We hope these findings can inform future efforts to better prepare foundation models for embodied intelligence.

%%%%%%%%% ICLR-required / recommended statements (do not count toward the page
%%%%%%%%% limit). Placed at the end of the main text, before the references.

\section*{Acknowledgments}

This work is supported by CMU Foundation and Language Model (FLAME) Center for computational resources. This work used Delta and DeltaAI at National Center for Supercomputing Applications through allocation CIS250941 from the Advanced Cyberinfrastructure Coordination Ecosystem: Services \& Support (ACCESS) program, which is supported by U.S. National Science Foundation grants \#2138259, \#2138286, \#2138307, \#2137603, and \#2138296. We sincerely appreciate Cathy Jiao, Zichun Yu, Shanshan Zhong, Xiaochuan Li and Hao Kang for providing helpful feedback on this work.

\section*{AI Use Statement}
In this work, we used generative AI tools including ChatGPT and Claude to polish the writing, and generate code to help implement the proposed method. We have not used generative AI for ideation, hypothesis formulation, dataset generation, or result interpretation. We have reviewed all the content generated by AI, including polished texts and codes. We take responsibility for the final content of this work, including text, claims or artifacts produced with the aid of generative AI.

% \section*{Ethics Statement}

\section*{Reproducibility Statement}
We provide the information needed to reproduce our results in the main text and
appendix. The proximity-based data engine and mid-training procedure are
described in Section~\ref{sec:midtrain}, and the full experimental setup is given
in Section~\ref{sec:setup}. Implementation details and hyperparameters for VLM
mid-training and VLA fine-tuning are reported in
Appendices~\ref{app:impl_detail_vlm},~\ref{app:impl_detail_vla}~and~\ref{app:impl_detail_real_world}. As stated in the abstract, we will release all
code, data, and models to support reproduction.

\bibliographystyle{iclr2027_conference}
\bibliography{iclr2027_conference}

@article{gretton2012kernel,
  author  = {Arthur Gretton and Karsten M. Borgwardt and Malte J. Rasch and Bernhard Sch{{\"o}}lkopf and Alexander Smola},
  title   = {A Kernel Two-Sample Test},
  journal = {Journal of Machine Learning Research},
  year    = {2012},
  volume  = {13},
  number  = {25},
  pages   = {723-773},
  url     = {http://jmlr.org/papers/v13/gretton12a.html}
}

@InProceedings{pmlr-v202-li23q,
  title = 	 {{BLIP}-2: Bootstrapping Language-Image Pre-training with Frozen Image Encoders and Large Language Models},
  author =       {Li, Junnan and Li, Dongxu and Savarese, Silvio and Hoi, Steven},
  booktitle = 	 {Proceedings of the 40th International Conference on Machine Learning},
  pages = 	 {19730--19742},
  year = 	 {2023},
  editor = 	 {Krause, Andreas and Brunskill, Emma and Cho, Kyunghyun and Engelhardt, Barbara and Sabato, Sivan and Scarlett, Jonathan},
  volume = 	 {202},
  series = 	 {Proceedings of Machine Learning Research},
  month = 	 {23--29 Jul},
  publisher =    {PMLR},
  url = 	 {https://proceedings.mlr.press/v202/li23q.html}
}

@inproceedings{llava,
 author = {Liu, Haotian and Li, Chunyuan and Wu, Qingyang and Lee, Yong Jae},
 booktitle = {Advances in Neural Information Processing Systems},
 doi = {10.52202/075280-1516},
 editor = {A. Oh and T. Naumann and A. Globerson and K. Saenko and M. Hardt and S. Levine},
 pages = {34892--34916},
 publisher = {Curran Associates, Inc.},
 title = {Visual Instruction Tuning},
 url = {https://proceedings.neurips.cc/paper_files/paper/2023/file/6dcf277ea32ce3288914faf369fe6de0-Paper-Conference.pdf},
 volume = {36},
 year = {2023}
}

@inproceedings{xing2025shortcut,
  title={Shortcut learning in generalist robot policies: The role of dataset diversity and fragmentation},
  author={Xing, Youguang and Luo, Xu and Xie, Junlin and Gao, Lianli and Shen, Heng Tao and Song, Jingkuan},
  booktitle={Conference on Robot Learning},
  pages={3239--3266},
  year={2025}
}

@article{zhang2026vlm4vla,
  title={VLM4VLA: Revisiting Vision-Language-Models in Vision-Language-Action Models},
  author={Zhang, Jianke and Chen, Xiaoyu and Wang, Qiuyue and Li, Mingsheng and Guo, Yanjiang and Hu, Yucheng and Zhang, Jiajun and Bai, Shuai and Lin, Junyang and Chen, Jianyu},
  journal={arXiv preprint arXiv:2601.03309},
  year={2026}
}

@article{yang2025vlaser,
  title={Vlaser: Vision-Language-Action Model with Synergistic Embodied Reasoning},
  author={Yang, Ganlin and Zhang, Tianyi and Hao, Haoran and Wang, Weiyun and Liu, Yibin and Wang, Dehui and Chen, Guanzhou and Cai, Zijian and Chen, Junting and Su, Weijie and others},
  journal={arXiv preprint arXiv:2510.11027},
  year={2025}
}

@InProceedings{li2022blip,
  title = 	 {{BLIP}: Bootstrapping Language-Image Pre-training for Unified Vision-Language Understanding and Generation},
  author =       {Li, Junnan and Li, Dongxu and Xiong, Caiming and Hoi, Steven},
  booktitle = 	 {Proceedings of the 39th International Conference on Machine Learning},
  pages = 	 {12888--12900},
  year = 	 {2022},
  editor = 	 {Chaudhuri, Kamalika and Jegelka, Stefanie and Song, Le and Szepesvari, Csaba and Niu, Gang and Sabato, Sivan},
  volume = 	 {162},
  series = 	 {Proceedings of Machine Learning Research},
  month = 	 {17--23 Jul},
  publisher =    {PMLR},
  url = 	 {https://proceedings.mlr.press/v162/li22n.html}
}

@misc{laion400m,
      title={{LAION-400M}: Open Dataset of CLIP-Filtered 400 Million Image-Text Pairs}, 
      author={Christoph Schuhmann and Richard Vencu and Romain Beaumont and Robert Kaczmarczyk and Clayton Mullis and Aarush Katta and Theo Coombes and Jenia Jitsev and Aran Komatsuzaki},
      year={2021},
      eprint={2111.02114},
      archivePrefix={arXiv},
      primaryClass={cs.CV},
      url={https://arxiv.org/abs/2111.02114}, 
}

@misc{liu2024improvedbaselinesvisualinstruction,
      title={Improved Baselines with Visual Instruction Tuning}, 
      author={Haotian Liu and Chunyuan Li and Yuheng Li and Yong Jae Lee},
      year={2024},
      eprint={2310.03744},
      archivePrefix={arXiv},
      primaryClass={cs.CV},
      url={https://arxiv.org/abs/2310.03744}, 
}

@inproceedings{changpinyo2021cc12m,
  title = {{Conceptual 12M}: Pushing Web-Scale Image-Text Pre-Training To Recognize Long-Tail Visual Concepts},
  author = {Changpinyo, Soravit and Sharma, Piyush and Ding, Nan and Soricut, Radu},
  booktitle = {CVPR},
  year = {2021},
}

@InProceedings{zellers2019vcr,
author = {Zellers, Rowan and Bisk, Yonatan and Farhadi, Ali and Choi, Yejin},
title = {From Recognition to Cognition: Visual Commonsense Reasoning},
booktitle = {Proceedings of the IEEE/CVF Conference on Computer Vision and Pattern Recognition (CVPR)},
month = {June},
year = {2019}
}

@article{zhou2025roborefer,
  title={RoboRefer: Towards Spatial Referring with Reasoning in Vision-Language Models for Robotics},
  author={Zhou, Enshen and An, Jingkun and Chi, Cheng and Han, Yi and Rong, Shanyu and Zhang, Chi and Wang, Pengwei and Wang, Zhongyuan and Huang, Tiejun and Sheng, Lu and others},
  journal={arXiv preprint arXiv:2506.04308},
  year={2025}
}

@inproceedings{du-etal-2024-embspatial,
    title = "{E}mb{S}patial-Bench: Benchmarking Spatial Understanding for Embodied Tasks with Large Vision-Language Models",
    author = "Du, Mengfei  and
      Wu, Binhao  and
      Li, Zejun  and
      Huang, Xuanjing  and
      Wei, Zhongyu",
    editor = "Ku, Lun-Wei  and
      Martins, Andre  and
      Srikumar, Vivek",
    booktitle = "Proceedings of the 62nd Annual Meeting of the Association for Computational Linguistics (Volume 2: Short Papers)",
    month = aug,
    year = "2024",
    address = "Bangkok, Thailand",
    publisher = "Association for Computational Linguistics",
    url = "https://aclanthology.org/2024.acl-short.33/",
    doi = "10.18653/v1/2024.acl-short.33",
    pages = "346--355"
}

@misc{chen2025robo2vlmvisualquestionanswering,
      title={Robo2VLM: Visual Question Answering from Large-Scale In-the-Wild Robot Manipulation Datasets}, 
      author={Kaiyuan Chen and Shuangyu Xie and Zehan Ma and Pannag R Sanketi and Ken Goldberg},
      year={2025},
      eprint={2505.15517},
      archivePrefix={arXiv},
      primaryClass={cs.RO},
      url={https://arxiv.org/abs/2505.15517}, 
}

@InProceedings{pmlr-v270-yuan25c-robopoint,
  title = 	 {RoboPoint: A Vision-Language Model for Spatial Affordance Prediction in Robotics},
  author =       {Yuan, Wentao and Duan, Jiafei and Blukis, Valts and Pumacay, Wilbert and Krishna, Ranjay and Murali, Adithyavairavan and Mousavian, Arsalan and Fox, Dieter},
  booktitle = 	 {Proceedings of The 8th Conference on Robot Learning},
  pages = 	 {4005--4020},
  year = 	 {2025},
  editor = 	 {Agrawal, Pulkit and Kroemer, Oliver and Burgard, Wolfram},
  volume = 	 {270},
  series = 	 {Proceedings of Machine Learning Research},
  month = 	 {06--09 Nov},
  publisher =    {PMLR},
  url = 	 {https://proceedings.mlr.press/v270/yuan25c.html}
}

@article{mees2022calvin,
author = {Oier Mees and Lukas Hermann and Erick Rosete-Beas and Wolfram Burgard},
title = {CALVIN: A Benchmark for Language-Conditioned Policy Learning for Long-Horizon Robot Manipulation Tasks},
journal={IEEE Robotics and Automation Letters (RA-L)},
volume={7},
number={3},
pages={7327-7334},
year={2022}
}

@article{li2025simpler,
         title={Evaluating Real-World Robot Manipulation Policies in Simulation},
         author={Xuanlin Li and Kyle Hsu and Jiayuan Gu and Karl Pertsch and Oier Mees and Homer Rich Walke and Chuyuan Fu and Ishikaa Lunawat and Isabel Sieh and Sean Kirmani and Sergey Levine and Jiajun Wu and Chelsea Finn and Hao Su and Quan Vuong and Ted Xiao},
         journal = {arXiv preprint arXiv:2405.05941},
         year={2024}
}

@article{liu2023libero,
  title={LIBERO: Benchmarking Knowledge Transfer for Lifelong Robot Learning},
  author={Liu, Bo and Zhu, Yifeng and Gao, Chongkai and Feng, Yihao and Liu, Qiang and Zhu, Yuke and Stone, Peter},
  journal={arXiv preprint arXiv:2306.03310},
  year={2023}
}

@inproceedings{walke2023bridgedata,
  title={BridgeData V2: A Dataset for Robot Learning at Scale},
  author={Walke, Homer and Black, Kevin and Lee, Abraham and Kim, Moo Jin and Du, Max and Zheng, Chongyi and Zhao, Tony and Hansen-Estruch, Philippe and Vuong, Quan and He, Andre and Myers, Vivek and Fang, Kuan and Finn, Chelsea and Levine, Sergey},
  booktitle={Conference on Robot Learning (CoRL)},
  year={2023}
}

@article{wang2025internvl3_5,
  title={InternVL3.5: Advancing Open-Source Multimodal Models in Versatility, Reasoning, and Efficiency},
  author={Wang, Weiyun and Gao, Zhangwei and Gu, Lixin and Pu, Hengjun and Cui, Long and Wei, Xingguang and Liu, Zhaoyang and Jing, Linglin and Ye, Shenglong and Shao, Jie and others},
  journal={arXiv preprint arXiv:2508.18265},
  year={2025}
}

@misc{Qwen-VL,
      title={Qwen-VL: A Versatile Vision-Language Model for Understanding, Localization, Text Reading, and Beyond}, 
      author={Jinze Bai and Shuai Bai and Shusheng Yang and Shijie Wang and Sinan Tan and Peng Wang and Junyang Lin and Chang Zhou and Jingren Zhou},
      year={2023},
      eprint={2308.12966},
      archivePrefix={arXiv},
      primaryClass={cs.CV},
      url={https://arxiv.org/abs/2308.12966}, 
}

@article{xie2023data,
  author = {Sang Michael Xie and Shibani Santurkar and Tengyu Ma and Percy Liang},
  journal = {Advances in Neural Information Processing Systems (NeurIPS)},
  title = {Data Selection for Language Models via Importance Resampling},
  year = {2023},
}

@inproceedings{zheng2024llamafactory,
  title={LlamaFactory: Unified Efficient Fine-Tuning of 100+ Language Models},
  author={Yaowei Zheng and Richong Zhang and Junhao Zhang and Yanhan Ye and Zheyan Luo and Zhangchi Feng and Yongqiang Ma},
  booktitle={Proceedings of the 62nd Annual Meeting of the Association for Computational Linguistics (Volume 3: System Demonstrations)},
  address={Bangkok, Thailand},
  publisher={Association for Computational Linguistics},
  year={2024},
  url={http://arxiv.org/abs/2403.13372}
}

@article{beyer2024paligemma,
  title        = {PaliGemma: A Versatile 3B VLM for Transfer},
  author       = {Beyer, Lucas and Steiner, Andreas and Pinto, Andr{\'e} Susano and Kolesnikov, Alexander and Wang, Xiao and Salz, Daniel and Neumann, Maxim and Alabdulmohsin, Ibrahim and Tschannen, Michael and Bugliarello, Emanuele and Unterthiner, Thomas and Keysers, Daniel and Koppula, Skanda and Liu, Fangyu and Grycner, Adam and Gritsenko, Alexey and Houlsby, Neil and Kumar, Manoj and Rong, Keran and Eisenschlos, Julian and Kabra, Rishabh and Bauer, Matthias and Bo{\v{s}}njak, Matko and Chen, Xi and Minderer, Matthias and Voigtlaender, Paul and Bica, Ioana and Balazevic, Ivana and Puigcerver, Joan and Papalampidi, Pinelopi and Henaff, Olivier and Xiong, Xi and Soricut, Radu and Harmsen, Jeremiah and Zhai, Xiaohua},
  journal      = {arXiv preprint arXiv:2407.07726},
  year         = {2024},
  eprint       = {2407.07726},
  archivePrefix= {arXiv},
  primaryClass = {cs.CV}
}

@article{steiner2024paligemma2,
  title        = {PaliGemma 2: A Family of Versatile VLMs for Transfer},
  author       = {Steiner, Andreas and Pinto, Andr{\'e} Susano and Tschannen, Michael and Keysers, Daniel and Wang, Xiao and Bitton, Yonatan and Gritsenko, Alexey and Minderer, Matthias and Sherbondy, Anthony and Long, Shangbang and Qin, Siyang and Ingle, Reeve and Bugliarello, Emanuele and Kazemzadeh, Sahar and Mesnard, Thomas and Alabdulmohsin, Ibrahim and Beyer, Lucas and Zhai, Xiaohua},
  journal      = {arXiv preprint arXiv:2412.03555},
  year         = {2024},
  eprint       = {2412.03555},
  archivePrefix= {arXiv},
  primaryClass = {cs.CV}
}

@article{peng2023kosmos2,
  title        = {Kosmos-2: Grounding Multimodal Large Language Models to the World},
  author       = {Peng, Zhiliang and Wang, Wenhui and Dong, Li and Hao, Yaru and Huang, Shaohan and Ma, Shuming and Wei, Furu},
  journal      = {arXiv preprint arXiv:2306.14824},
  year         = {2023},
  eprint       = {2306.14824},
  archivePrefix= {arXiv},
  primaryClass = {cs.CV}
}

@article{kim2024openvla,
  title         = {OpenVLA: An Open-Source Vision-Language-Action Model},
  author        = {Kim, Moo Jin and Pertsch, Karl and Karamcheti, Siddharth and Xiao, Ted and Balakrishna, Ashwin and Nair, Suraj and Rafailov, Rafael and Foster, Ethan and Lam, Grace and Sanketi, Pannag and Vuong, Quan and Kollar, Thomas and Burchfiel, Benjamin and Tedrake, Russ and Sadigh, Dorsa and Levine, Sergey and Liang, Percy and Finn, Chelsea},
  journal       = {arXiv preprint arXiv:2406.09246},
  year          = {2024},
  eprint        = {2406.09246},
  archivePrefix = {arXiv},
  primaryClass  = {cs.RO}
}

@article{black2024pi0,
  title         = {$\pi_0$: A Vision-Language-Action Flow Model for General Robot Control},
  author        = {Black, Kevin and Brown, Noah and Driess, Danny and Esmail, Adnan and Equi, Michael and Finn, Chelsea and Fusai, Niccolo and Groom, Lachy and Hausman, Karol and Ichter, Brian and Jakubczak, Szymon and Jones, Tim and Ke, Liyiming and Levine, Sergey and Li-Bell, Adrian and Mothukuri, Mohith and Nair, Suraj and Pertsch, Karl and Shi, Lucy Xiaoyang and Tanner, James and Vuong, Quan and Walling, Anna and Wang, Haohuan and Zhilinsky, Ury},
  journal       = {arXiv preprint arXiv:2410.24164},
  year          = {2024},
  eprint        = {2410.24164},
  archivePrefix = {arXiv},
  primaryClass  = {cs.LG}
}

@misc{gemini_robotics_2025,
      title={Gemini Robotics 1.5: Pushing the Frontier of Generalist Robots with Advanced Embodied Reasoning, Thinking, and Motion Transfer}, 
      author={{Gemini Robotics Team} and Abbas Abdolmaleki and Saminda Abeyruwan and Joshua Ainslie and Jean-Baptiste Alayrac and Montserrat Gonzalez Arenas and Ashwin Balakrishna and Nathan Batchelor and Alex Bewley and Jeff Bingham and Michael Bloesch and Konstantinos Bousmalis and Philemon Brakel and Anthony Brohan and Thomas Buschmann and Arunkumar Byravan and Serkan Cabi and Ken Caluwaerts and Federico Casarini and Christine Chan and Oscar Chang and London Chappellet-Volpini and Jose Enrique Chen and Xi Chen and Hao-Tien Lewis Chiang and Krzysztof Choromanski and Adrian Collister and David B. D'Ambrosio and Sudeep Dasari and Todor Davchev and Meet Kirankumar Dave and Coline Devin and Norman Di Palo and Tianli Ding and Carl Doersch and Adil Dostmohamed and Yilun Du and Debidatta Dwibedi and Sathish Thoppay Egambaram and Michael Elabd and Tom Erez and Xiaolin Fang and Claudio Fantacci and Cody Fong and Erik Frey and Chuyuan Fu and Ruiqi Gao and Marissa Giustina and Keerthana Gopalakrishnan and Laura Graesser and Oliver Groth and Agrim Gupta and Roland Hafner and Steven Hansen and Leonard Hasenclever and Sam Haves and Nicolas Heess and Brandon Hernaez and Alex Hofer and Jasmine Hsu and Lu Huang and Sandy H. Huang and Atil Iscen and Mithun George Jacob and Deepali Jain and Sally Jesmonth and Abhishek Jindal and Ryan Julian and Dmitry Kalashnikov and M. Emre Karagozler and Stefani Karp and Matija Kecman and J. Chase Kew and Donnie Kim and Frank Kim and Junkyung Kim and Thomas Kipf and Sean Kirmani and Ksenia Konyushkova and Li Yang Ku and Yuheng Kuang and Thomas Lampe and Antoine Laurens and Tuan Anh Le and Isabel Leal and Alex X. Lee and Tsang-Wei Edward Lee and Guy Lever and Jacky Liang and Li-Heng Lin and Fangchen Liu and Shangbang Long and Caden Lu and Sharath Maddineni and Anirudha Majumdar and Kevis-Kokitsi Maninis and Andrew Marmon and Sergio Martinez and Assaf Hurwitz Michaely and Niko Milonopoulos and Joss Moore and Robert Moreno and Michael Neunert and Francesco Nori and Joy Ortiz and Kenneth Oslund and Carolina Parada and Emilio Parisotto and Amaris Paryag and Acorn Pooley and Thomas Power and Alessio Quaglino and Haroon Qureshi and Rajkumar Vasudeva Raju and Helen Ran and Dushyant Rao and Kanishka Rao and Isaac Reid and David Rendleman and Krista Reymann and Miguel Rivas and Francesco Romano and Yulia Rubanova and Peter Pastor Sampedro and Pannag R Sanketi and Dhruv Shah and Mohit Sharma and Kathryn Shea and Mohit Shridhar and Charles Shu and Vikas Sindhwani and Sumeet Singh and Radu Soricut and Rachel Sterneck and Ian Storz and Razvan Surdulescu and Jie Tan and Jonathan Tompson and Saran Tunyasuvunakool and Jake Varley and Grace Vesom and Giulia Vezzani and Maria Bauza Villalonga and Oriol Vinyals and René Wagner and Ayzaan Wahid and Stefan Welker and Paul Wohlhart and Chengda Wu and Markus Wulfmeier and Fei Xia and Ted Xiao and Annie Xie and Jinyu Xie and Peng Xu and Sichun Xu and Ying Xu and Zhuo Xu and Jimmy Yan and Sherry Yang and Skye Yang and Yuxiang Yang and Hiu Hong Yu and Wenhao Yu and Wentao Yuan and Yuan Yuan and Jingwei Zhang and Tingnan Zhang and Zhiyuan Zhang and Allan Zhou and Guangyao Zhou and Yuxiang Zhou},
      year={2025},
      eprint={2510.03342},
      archivePrefix={arXiv},
      primaryClass={cs.RO},
      url={https://arxiv.org/abs/2510.03342}, 
}

@article{ji2025robobrain,
  title   = {RoboBrain: A Unified Brain Model for Robotic Manipulation from Abstract to Concrete},
  author  = {Ji, Yuheng and others},
  journal = {arXiv preprint arXiv:2502.21257},
  year    = {2025}
}

@INPROCEEDINGS{kim2025oft, 
    AUTHOR    = {Moo Jin Kim AND Chelsea Finn AND Percy Liang}, 
    TITLE     = {{Fine-Tuning Vision-Language-Action Models: Optimizing Speed and Success}}, 
    BOOKTITLE = {Proceedings of Robotics: Science and Systems}, 
    YEAR      = {2025}, 
    ADDRESS   = {LosAngeles, CA, USA}, 
    MONTH     = {June}, 
    DOI       = {10.15607/RSS.2025.XXI.017} 
}

@misc{li2025eagle2,
      title={{Eagle 2}: Building Post-Training Data Strategies from Scratch for Frontier Vision-Language Models}, 
      author={Zhiqi Li and Guo Chen and Shilong Liu and Shihao Wang and Vibashan VS and Yishen Ji and Shiyi Lan and Hao Zhang and Yilin Zhao and Subhashree Radhakrishnan and Nadine Chang and Karan Sapra and Amala Sanjay Deshmukh and Tuomas Rintamaki and Matthieu Le and Ilia Karmanov and Lukas Voegtle and Philipp Fischer and De-An Huang and Timo Roman and Tong Lu and Jose M. Alvarez and Bryan Catanzaro and Jan Kautz and Andrew Tao and Guilin Liu and Zhiding Yu},
      year={2025},
      eprint={2501.14818},
      archivePrefix={arXiv},
      primaryClass={cs.CV},
      url={https://arxiv.org/abs/2501.14818}, 
}

@misc{gr00tn1_2025,
      title={GR00T N1: An Open Foundation Model for Generalist Humanoid Robots}, 
      author={NVIDIA and Johan Bjorck and Fernando Castañeda and Nikita Cherniadev and Xingye Da and Runyu Ding and Linxi "Jim" Fan and Yu Fang and Dieter Fox and Fengyuan Hu and Spencer Huang and Joel Jang and Zhenyu Jiang and Jan Kautz and Kaushil Kundalia and Lawrence Lao and Zhiqi Li and Zongyu Lin and Kevin Lin and Guilin Liu and Edith Llontop and Loic Magne and Ajay Mandlekar and Avnish Narayan and Soroush Nasiriany and Scott Reed and You Liang Tan and Guanzhi Wang and Zu Wang and Jing Wang and Qi Wang and Jiannan Xiang and Yuqi Xie and Yinzhen Xu and Zhenjia Xu and Seonghyeon Ye and Zhiding Yu and Ao Zhang and Hao Zhang and Yizhou Zhao and Ruijie Zheng and Yuke Zhu},
      year={2025},
      eprint={2503.14734},
      archivePrefix={arXiv},
      primaryClass={cs.RO},
      url={https://arxiv.org/abs/2503.14734}, 
}

@misc{wang2025octothinker,
      title={OctoThinker: Mid-training Incentivizes Reinforcement Learning Scaling}, 
      author={Zengzhi Wang and Fan Zhou and Xuefeng Li and Pengfei Liu},
      year={2025},
      eprint={2506.20512},
      archivePrefix={arXiv},
      primaryClass={cs.CL},
      url={https://arxiv.org/abs/2506.20512}, 
}

@article{Qwen3-VL,
      title={Qwen3-VL Technical Report}, 
      author={Shuai Bai and Yuxuan Cai and Ruizhe Chen and Keqin Chen and Xionghui Chen and Zesen Cheng and Lianghao Deng and Wei Ding and Chang Gao and Chunjiang Ge and Wenbin Ge and Zhifang Guo and Qidong Huang and Jie Huang and Fei Huang and Binyuan Hui and Shutong Jiang and Zhaohai Li and Mingsheng Li and Mei Li and Kaixin Li and Zicheng Lin and Junyang Lin and Xuejing Liu and Jiawei Liu and Chenglong Liu and Yang Liu and Dayiheng Liu and Shixuan Liu and Dunjie Lu and Ruilin Luo and Chenxu Lv and Rui Men and Lingchen Meng and Xuancheng Ren and Xingzhang Ren and Sibo Song and Yuchong Sun and Jun Tang and Jianhong Tu and Jianqiang Wan and Peng Wang and Pengfei Wang and Qiuyue Wang and Yuxuan Wang and Tianbao Xie and Yiheng Xu and Haiyang Xu and Jin Xu and Zhibo Yang and Mingkun Yang and Jianxin Yang and An Yang and Bowen Yu and Fei Zhang and Hang Zhang and Xi Zhang and Bo Zheng and Humen Zhong and Jingren Zhou and Fan Zhou and Jing Zhou and Yuanzhi Zhu and Ke Zhu},
	  journal={arXiv preprint arXiv:2511.21631},
      year={2025}
}

@article{Qwen2.5-VL,
  title={Qwen2.5-VL Technical Report},
  author={Bai, Shuai and Chen, Keqin and Liu, Xuejing and Wang, Jialin and Ge, Wenbin and Song, Sibo and Dang, Kai and Wang, Peng and Wang, Shijie and Tang, Jun and Zhong, Humen and Zhu, Yuanzhi and Yang, Mingkun and Li, Zhaohai and Wan, Jianqiang and Wang, Pengfei and Ding, Wei and Fu, Zheren and Xu, Yiheng and Ye, Jiabo and Zhang, Xi and Xie, Tianbao and Cheng, Zesen and Zhang, Hang and Yang, Zhibo and Xu, Haiyang and Lin, Junyang},
  journal={arXiv preprint arXiv:2502.13923},
  year={2025}
}

@article{Qwen2-VL,
  title={Qwen2-VL: Enhancing Vision-Language Model's Perception of the World at Any Resolution},
  author={Wang, Peng and Bai, Shuai and Tan, Sinan and Wang, Shijie and Fan, Zhihao and Bai, Jinze and Chen, Keqin and Liu, Xuejing and Wang, Jialin and Ge, Wenbin and Fan, Yang and Dang, Kai and Du, Mengfei and Ren, Xuancheng and Men, Rui and Liu, Dayiheng and Zhou, Chang and Zhou, Jingren and Lin, Junyang},
  journal={arXiv preprint arXiv:2409.12191},
  year={2024}
}

@misc{internvl25,
      title={Expanding Performance Boundaries of Open-Source Multimodal Models with Model, Data, and Test-Time Scaling}, 
      author={Zhe Chen and Weiyun Wang and Yue Cao and Yangzhou Liu and Zhangwei Gao and Erfei Cui and Jinguo Zhu and Shenglong Ye and Hao Tian and Zhaoyang Liu and Lixin Gu and Xuehui Wang and Qingyun Li and Yiming Ren and Zixuan Chen and Jiapeng Luo and Jiahao Wang and Tan Jiang and Bo Wang and Conghui He and Botian Shi and Xingcheng Zhang and Han Lv and Yi Wang and Wenqi Shao and Pei Chu and Zhongying Tu and Tong He and Zhiyong Wu and Huipeng Deng and Jiaye Ge and Kai Chen and Kaipeng Zhang and Limin Wang and Min Dou and Lewei Lu and Xizhou Zhu and Tong Lu and Dahua Lin and Yu Qiao and Jifeng Dai and Wenhai Wang},
      year={2025},
      eprint={2412.05271},
      archivePrefix={arXiv},
      primaryClass={cs.CV},
      url={https://arxiv.org/abs/2412.05271}, 
}

@inproceedings{cai2025spatialbot,
  title={Spatialbot: Precise spatial understanding with vision language models},
  author={Cai, Wenxiao and Ponomarenko, Iaroslav and Yuan, Jianhao and Li, Xiaoqi and Yang, Wankou and Dong, Hao and Zhao, Bo},
  booktitle={2025 IEEE International Conference on Robotics and Automation (ICRA)},
  pages={9490--9498},
  year={2025},
  organization={IEEE}
}

@misc{olmo20252olmo2furious,
      title={2 OLMo 2 Furious}, 
      author={{Team OLMo} and Pete Walsh and Luca Soldaini and Dirk Groeneveld and Kyle Lo and Shane Arora and Akshita Bhagia and Yuling Gu and Shengyi Huang and Matt Jordan and Nathan Lambert and Dustin Schwenk and Oyvind Tafjord and Taira Anderson and David Atkinson and Faeze Brahman and Christopher Clark and Pradeep Dasigi and Nouha Dziri and Allyson Ettinger and Michal Guerquin and David Heineman and Hamish Ivison and Pang Wei Koh and Jiacheng Liu and Saumya Malik and William Merrill and Lester James V. Miranda and Jacob Morrison and Tyler Murray and Crystal Nam and Jake Poznanski and Valentina Pyatkin and Aman Rangapur and Michael Schmitz and Sam Skjonsberg and David Wadden and Christopher Wilhelm and Michael Wilson and Luke Zettlemoyer and Ali Farhadi and Noah A. Smith and Hannaneh Hajishirzi},
      year={2025},
      eprint={2501.00656},
      archivePrefix={arXiv},
      primaryClass={cs.CL},
      url={https://arxiv.org/abs/2501.00656}, 
}

@misc{grattafiori2024llama3herdmodels,
      title={The Llama 3 Herd of Models}, 
      author={Aaron Grattafiori and Abhimanyu Dubey and Abhinav Jauhri and Abhinav Pandey and Abhishek Kadian and Ahmad Al-Dahle and Aiesha Letman and Akhil Mathur and Alan Schelten and Alex Vaughan and Amy Yang and Angela Fan and Anirudh Goyal and Anthony Hartshorn and Aobo Yang and Archi Mitra and Archie Sravankumar and Artem Korenev and Arthur Hinsvark and Arun Rao and Aston Zhang and Aurelien Rodriguez and Austen Gregerson and Ava Spataru and Baptiste Roziere and Bethany Biron and Binh Tang and Bobbie Chern and Charlotte Caucheteux and Chaya Nayak and Chloe Bi and Chris Marra and Chris McConnell and Christian Keller and Christophe Touret and Chunyang Wu and Corinne Wong and Cristian Canton Ferrer and Cyrus Nikolaidis and Damien Allonsius and Daniel Song and Danielle Pintz and Danny Livshits and Danny Wyatt and David Esiobu and Dhruv Choudhary and Dhruv Mahajan and Diego Garcia-Olano and Diego Perino and Dieuwke Hupkes and Egor Lakomkin and Ehab AlBadawy and Elina Lobanova and Emily Dinan and Eric Michael Smith and Filip Radenovic and Francisco Guzmán and Frank Zhang and Gabriel Synnaeve and Gabrielle Lee and Georgia Lewis Anderson and Govind Thattai and Graeme Nail and Gregoire Mialon and Guan Pang and Guillem Cucurell and Hailey Nguyen and Hannah Korevaar and Hu Xu and Hugo Touvron and Iliyan Zarov and Imanol Arrieta Ibarra and Isabel Kloumann and Ishan Misra and Ivan Evtimov and Jack Zhang and Jade Copet and Jaewon Lee and Jan Geffert and Jana Vranes and Jason Park and Jay Mahadeokar and Jeet Shah and Jelmer van der Linde and Jennifer Billock and Jenny Hong and Jenya Lee and Jeremy Fu and Jianfeng Chi and Jianyu Huang and Jiawen Liu and Jie Wang and Jiecao Yu and Joanna Bitton and Joe Spisak and Jongsoo Park and Joseph Rocca and Joshua Johnstun and Joshua Saxe and Junteng Jia and Kalyan Vasuden Alwala and Karthik Prasad and Kartikeya Upasani and Kate Plawiak and Ke Li and Kenneth Heafield and Kevin Stone and Khalid El-Arini and Krithika Iyer and Kshitiz Malik and Kuenley Chiu and Kunal Bhalla and Kushal Lakhotia and Lauren Rantala-Yeary and Laurens van der Maaten and Lawrence Chen and Liang Tan and Liz Jenkins and Louis Martin and Lovish Madaan and Lubo Malo and Lukas Blecher and Lukas Landzaat and Luke de Oliveira and Madeline Muzzi and Mahesh Pasupuleti and Mannat Singh and Manohar Paluri and Marcin Kardas and Maria Tsimpoukelli and Mathew Oldham and Mathieu Rita and Maya Pavlova and Melanie Kambadur and Mike Lewis and Min Si and Mitesh Kumar Singh and Mona Hassan and Naman Goyal and Narjes Torabi and Nikolay Bashlykov and Nikolay Bogoychev and Niladri Chatterji and Ning Zhang and Olivier Duchenne and Onur Çelebi and Patrick Alrassy and Pengchuan Zhang and Pengwei Li and Petar Vasic and Peter Weng and Prajjwal Bhargava and Pratik Dubal and Praveen Krishnan and Punit Singh Koura and Puxin Xu and Qing He and Qingxiao Dong and Ragavan Srinivasan and Raj Ganapathy and Ramon Calderer and Ricardo Silveira Cabral and Robert Stojnic and Roberta Raileanu and Rohan Maheswari and Rohit Girdhar and Rohit Patel and Romain Sauvestre and Ronnie Polidoro and Roshan Sumbaly and Ross Taylor and Ruan Silva and Rui Hou and Rui Wang and Saghar Hosseini and Sahana Chennabasappa and Sanjay Singh and Sean Bell and Seohyun Sonia Kim and Sergey Edunov and Shaoliang Nie and Sharan Narang and Sharath Raparthy and Sheng Shen and Shengye Wan and Shruti Bhosale and Shun Zhang and Simon Vandenhende and Soumya Batra and Spencer Whitman and Sten Sootla and Stephane Collot and Suchin Gururangan and Sydney Borodinsky and Tamar Herman and Tara Fowler and Tarek Sheasha and Thomas Georgiou and Thomas Scialom and Tobias Speckbacher and Todor Mihaylov and Tong Xiao and Ujjwal Karn and Vedanuj Goswami and Vibhor Gupta and Vignesh Ramanathan and Viktor Kerkez and Vincent Gonguet and Virginie Do and Vish Vogeti and Vítor Albiero and Vladan Petrovic and Weiwei Chu and Wenhan Xiong and Wenyin Fu and Whitney Meers and Xavier Martinet and Xiaodong Wang and Xiaofang Wang and Xiaoqing Ellen Tan and Xide Xia and Xinfeng Xie and Xuchao Jia and Xuewei Wang and Yaelle Goldschlag and Yashesh Gaur and Yasmine Babaei and Yi Wen and Yiwen Song and Yuchen Zhang and Yue Li and Yuning Mao and Zacharie Delpierre Coudert and Zheng Yan and Zhengxing Chen and Zoe Papakipos and Aaditya Singh and Aayushi Srivastava and Abha Jain and Adam Kelsey and Adam Shajnfeld and Adithya Gangidi and Adolfo Victoria and Ahuva Goldstand and Ajay Menon and Ajay Sharma and Alex Boesenberg and Alexei Baevski and Allie Feinstein and Amanda Kallet and Amit Sangani and Amos Teo and Anam Yunus and Andrei Lupu and Andres Alvarado and Andrew Caples and Andrew Gu and Andrew Ho and Andrew Poulton and Andrew Ryan and Ankit Ramchandani and Annie Dong and Annie Franco and Anuj Goyal and Aparajita Saraf and Arkabandhu Chowdhury and Ashley Gabriel and Ashwin Bharambe and Assaf Eisenman and Azadeh Yazdan and Beau James and Ben Maurer and Benjamin Leonhardi and Bernie Huang and Beth Loyd and Beto De Paola and Bhargavi Paranjape and Bing Liu and Bo Wu and Boyu Ni and Braden Hancock and Bram Wasti and Brandon Spence and Brani Stojkovic and Brian Gamido and Britt Montalvo and Carl Parker and Carly Burton and Catalina Mejia and Ce Liu and Changhan Wang and Changkyu Kim and Chao Zhou and Chester Hu and Ching-Hsiang Chu and Chris Cai and Chris Tindal and Christoph Feichtenhofer and Cynthia Gao and Damon Civin and Dana Beaty and Daniel Kreymer and Daniel Li and David Adkins and David Xu and Davide Testuggine and Delia David and Devi Parikh and Diana Liskovich and Didem Foss and Dingkang Wang and Duc Le and Dustin Holland and Edward Dowling and Eissa Jamil and Elaine Montgomery and Eleonora Presani and Emily Hahn and Emily Wood and Eric-Tuan Le and Erik Brinkman and Esteban Arcaute and Evan Dunbar and Evan Smothers and Fei Sun and Felix Kreuk and Feng Tian and Filippos Kokkinos and Firat Ozgenel and Francesco Caggioni and Frank Kanayet and Frank Seide and Gabriela Medina Florez and Gabriella Schwarz and Gada Badeer and Georgia Swee and Gil Halpern and Grant Herman and Grigory Sizov and Guangyi and Zhang and Guna Lakshminarayanan and Hakan Inan and Hamid Shojanazeri and Han Zou and Hannah Wang and Hanwen Zha and Haroun Habeeb and Harrison Rudolph and Helen Suk and Henry Aspegren and Hunter Goldman and Hongyuan Zhan and Ibrahim Damlaj and Igor Molybog and Igor Tufanov and Ilias Leontiadis and Irina-Elena Veliche and Itai Gat and Jake Weissman and James Geboski and James Kohli and Janice Lam and Japhet Asher and Jean-Baptiste Gaya and Jeff Marcus and Jeff Tang and Jennifer Chan and Jenny Zhen and Jeremy Reizenstein and Jeremy Teboul and Jessica Zhong and Jian Jin and Jingyi Yang and Joe Cummings and Jon Carvill and Jon Shepard and Jonathan McPhie and Jonathan Torres and Josh Ginsburg and Junjie Wang and Kai Wu and Kam Hou U and Karan Saxena and Kartikay Khandelwal and Katayoun Zand and Kathy Matosich and Kaushik Veeraraghavan and Kelly Michelena and Keqian Li and Kiran Jagadeesh and Kun Huang and Kunal Chawla and Kyle Huang and Lailin Chen and Lakshya Garg and Lavender A and Leandro Silva and Lee Bell and Lei Zhang and Liangpeng Guo and Licheng Yu and Liron Moshkovich and Luca Wehrstedt and Madian Khabsa and Manav Avalani and Manish Bhatt and Martynas Mankus and Matan Hasson and Matthew Lennie and Matthias Reso and Maxim Groshev and Maxim Naumov and Maya Lathi and Meghan Keneally and Miao Liu and Michael L. Seltzer and Michal Valko and Michelle Restrepo and Mihir Patel and Mik Vyatskov and Mikayel Samvelyan and Mike Clark and Mike Macey and Mike Wang and Miquel Jubert Hermoso and Mo Metanat and Mohammad Rastegari and Munish Bansal and Nandhini Santhanam and Natascha Parks and Natasha White and Navyata Bawa and Nayan Singhal and Nick Egebo and Nicolas Usunier and Nikhil Mehta and Nikolay Pavlovich Laptev and Ning Dong and Norman Cheng and Oleg Chernoguz and Olivia Hart and Omkar Salpekar and Ozlem Kalinli and Parkin Kent and Parth Parekh and Paul Saab and Pavan Balaji and Pedro Rittner and Philip Bontrager and Pierre Roux and Piotr Dollar and Polina Zvyagina and Prashant Ratanchandani and Pritish Yuvraj and Qian Liang and Rachad Alao and Rachel Rodriguez and Rafi Ayub and Raghotham Murthy and Raghu Nayani and Rahul Mitra and Rangaprabhu Parthasarathy and Raymond Li and Rebekkah Hogan and Robin Battey and Rocky Wang and Russ Howes and Ruty Rinott and Sachin Mehta and Sachin Siby and Sai Jayesh Bondu and Samyak Datta and Sara Chugh and Sara Hunt and Sargun Dhillon and Sasha Sidorov and Satadru Pan and Saurabh Mahajan and Saurabh Verma and Seiji Yamamoto and Sharadh Ramaswamy and Shaun Lindsay and Shaun Lindsay and Sheng Feng and Shenghao Lin and Shengxin Cindy Zha and Shishir Patil and Shiva Shankar and Shuqiang Zhang and Shuqiang Zhang and Sinong Wang and Sneha Agarwal and Soji Sajuyigbe and Soumith Chintala and Stephanie Max and Stephen Chen and Steve Kehoe and Steve Satterfield and Sudarshan Govindaprasad and Sumit Gupta and Summer Deng and Sungmin Cho and Sunny Virk and Suraj Subramanian and Sy Choudhury and Sydney Goldman and Tal Remez and Tamar Glaser and Tamara Best and Thilo Koehler and Thomas Robinson and Tianhe Li and Tianjun Zhang and Tim Matthews and Timothy Chou and Tzook Shaked and Varun Vontimitta and Victoria Ajayi and Victoria Montanez and Vijai Mohan and Vinay Satish Kumar and Vishal Mangla and Vlad Ionescu and Vlad Poenaru and Vlad Tiberiu Mihailescu and Vladimir Ivanov and Wei Li and Wenchen Wang and Wenwen Jiang and Wes Bouaziz and Will Constable and Xiaocheng Tang and Xiaojian Wu and Xiaolan Wang and Xilun Wu and Xinbo Gao and Yaniv Kleinman and Yanjun Chen and Ye Hu and Ye Jia and Ye Qi and Yenda Li and Yilin Zhang and Ying Zhang and Yossi Adi and Youngjin Nam and Yu and Wang and Yu Zhao and Yuchen Hao and Yundi Qian and Yunlu Li and Yuzi He and Zach Rait and Zachary DeVito and Zef Rosnbrick and Zhaoduo Wen and Zhenyu Yang and Zhiwei Zhao and Zhiyu Ma},
      year={2024},
      eprint={2407.21783},
      archivePrefix={arXiv},
      primaryClass={cs.AI},
      url={https://arxiv.org/abs/2407.21783}, 
}

@misc{hu2024minicpmunveilingpotentialsmall,
      title={MiniCPM: Unveiling the Potential of Small Language Models with Scalable Training Strategies}, 
      author={Shengding Hu and Yuge Tu and Xu Han and Chaoqun He and Ganqu Cui and Xiang Long and Zhi Zheng and Yewei Fang and Yuxiang Huang and Weilin Zhao and Xinrong Zhang and Zheng Leng Thai and Kaihuo Zhang and Chongyi Wang and Yuan Yao and Chenyang Zhao and Jie Zhou and Jie Cai and Zhongwu Zhai and Ning Ding and Chao Jia and Guoyang Zeng and Dahai Li and Zhiyuan Liu and Maosong Sun},
      year={2024},
      eprint={2404.06395},
      archivePrefix={arXiv},
      primaryClass={cs.CL},
      url={https://arxiv.org/abs/2404.06395}, 
}

@inproceedings{GAN,
author = {Goodfellow, Ian J. and Pouget-Abadie, Jean and Mirza, Mehdi and Xu, Bing and Warde-Farley, David and Ozair, Sherjil and Courville, Aaron and Bengio, Yoshua},
title = {Generative adversarial nets},
year = {2014},
publisher = {MIT Press},
address = {Cambridge, MA, USA},
booktitle = {Proceedings of the 28th International Conference on Neural Information Processing Systems - Volume 2},
pages = {2672–2680},
numpages = {9},
location = {Montreal, Canada},
series = {NIPS'14}
}

@misc{RSCL,
      title={Contrastive Representation Regularization for Vision-Language-Action Models}, 
      author={Taeyoung Kim and Jimin Lee and Myungkyu Koo and Dongyoung Kim and Kyungmin Lee and Changyeon Kim and Younggyo Seo and Jinwoo Shin},
      year={2025},
      eprint={2510.01711},
      archivePrefix={arXiv},
      primaryClass={cs.RO},
      url={https://arxiv.org/abs/2510.01711}, 
}

@misc{internvlam1,
      title={InternVLA-M1: A Spatially Guided Vision-Language-Action Framework for Generalist Robot Policy}, 
      author={Xinyi Chen and Yilun Chen and Yanwei Fu and Ning Gao and Jiaya Jia and Weiyang Jin and Hao Li and Yao Mu and Jiangmiao Pang and Yu Qiao and Yang Tian and Bin Wang and Bolun Wang and Fangjing Wang and Hanqing Wang and Tai Wang and Ziqin Wang and Xueyuan Wei and Chao Wu and Shuai Yang and Jinhui Ye and Junqiu Yu and Jia Zeng and Jingjing Zhang and Jinyu Zhang and Shi Zhang and Feng Zheng and Bowen Zhou and Yangkun Zhu},
      year={2025},
      eprint={2510.13778},
      archivePrefix={arXiv},
      primaryClass={cs.RO},
      url={https://arxiv.org/abs/2510.13778}, 
}

@inproceedings{diversity,
author = {Wang, Tongzhou and Isola, Phillip},
title = {Understanding contrastive representation learning through alignment and uniformity on the hypersphere},
year = {2020},
publisher = {JMLR.org},
booktitle = {Proceedings of the 37th International Conference on Machine Learning},
articleno = {921},
numpages = {11},
series = {ICML'20}
}

@misc{fu2024blinkmultimodallargelanguage,
      title={BLINK: Multimodal Large Language Models Can See but Not Perceive}, 
      author={Xingyu Fu and Yushi Hu and Bangzheng Li and Yu Feng and Haoyu Wang and Xudong Lin and Dan Roth and Noah A. Smith and Wei-Chiu Ma and Ranjay Krishna},
      year={2024},
      eprint={2404.12390},
      archivePrefix={arXiv},
      primaryClass={cs.CV},
      url={https://arxiv.org/abs/2404.12390}, 
}

@inproceedings{POPE,
  title={Evaluating Object Hallucination in Large Vision-Language Models},
  author={Li, Yifan and Du, Yifan and Zhou, Kun and Wang, Jinpeng and Zhao, Wayne Xin and Wen, Ji-Rong},
  booktitle={The 2023 Conference on Empirical Methods in Natural Language Processing},
  year={2023},
  url={https://openreview.net/forum?id=xozJw0kZXF}
}

@misc{xu2025visulogic,
      title={VisuLogic: A Benchmark for Evaluating Visual Reasoning in Multi-modal Large Language Models}, 
      author={Weiye Xu and Jiahao Wang and Weiyun Wang and Zhe Chen and Wengang Zhou and Aijun Yang and Lewei Lu and Houqiang Li and Xiaohua Wang and Xizhou Zhu and Wenhai Wang and Jifeng Dai and Jinguo Zhu},
      year={2025},
      eprint={2504.15279},
      archivePrefix={arXiv},
      primaryClass={cs.CV},
      url={https://arxiv.org/abs/2504.15279}, 
}

@InProceedings{3DSRBench,
    author    = {Ma, Wufei and Chen, Haoyu and Zhang, Guofeng and Chou, Yu-Cheng and Chen, Jieneng and de Melo, Celso and Yuille, Alan},
    title     = {3DSRBench: A Comprehensive 3D Spatial Reasoning Benchmark},
    booktitle = {Proceedings of the IEEE/CVF International Conference on Computer Vision (ICCV)},
    month     = {October},
    year      = {2025},
    pages     = {6924-6934}
}

@inproceedings{spatialeval,
        title={Is A Picture Worth A Thousand Words? Delving Into Spatial Reasoning for Vision Language Models},
        author={Wang, Jiayu and Ming, Yifei and Shi, Zhenmei and Vineet, Vibhav and Wang, Xin and Li, Yixuan and Joshi, Neel},
        booktitle={The Thirty-Eighth Annual Conference on Neural Information Processing Systems},
        year={2024}
      }

@misc{khazatsky2025droidlargescaleinthewildrobot,
      title={DROID: A Large-Scale In-The-Wild Robot Manipulation Dataset}, 
      author={Alexander Khazatsky and Karl Pertsch and Suraj Nair and Ashwin Balakrishna and Sudeep Dasari and Siddharth Karamcheti and Soroush Nasiriany and Mohan Kumar Srirama and Lawrence Yunliang Chen and Kirsty Ellis and Peter David Fagan and Joey Hejna and Masha Itkina and Marion Lepert and Yecheng Jason Ma and Patrick Tree Miller and Jimmy Wu and Suneel Belkhale and Shivin Dass and Huy Ha and Arhan Jain and Abraham Lee and Youngwoon Lee and Marius Memmel and Sungjae Park and Ilija Radosavovic and Kaiyuan Wang and Albert Zhan and Kevin Black and Cheng Chi and Kyle Beltran Hatch and Shan Lin and Jingpei Lu and Jean Mercat and Abdul Rehman and Pannag R Sanketi and Archit Sharma and Cody Simpson and Quan Vuong and Homer Rich Walke and Blake Wulfe and Ted Xiao and Jonathan Heewon Yang and Arefeh Yavary and Tony Z. Zhao and Christopher Agia and Rohan Baijal and Mateo Guaman Castro and Daphne Chen and Qiuyu Chen and Trinity Chung and Jaimyn Drake and Ethan Paul Foster and Jensen Gao and Vitor Guizilini and David Antonio Herrera and Minho Heo and Kyle Hsu and Jiaheng Hu and Muhammad Zubair Irshad and Donovon Jackson and Charlotte Le and Yunshuang Li and Kevin Lin and Roy Lin and Zehan Ma and Abhiram Maddukuri and Suvir Mirchandani and Daniel Morton and Tony Nguyen and Abigail O'Neill and Rosario Scalise and Derick Seale and Victor Son and Stephen Tian and Emi Tran and Andrew E. Wang and Yilin Wu and Annie Xie and Jingyun Yang and Patrick Yin and Yunchu Zhang and Osbert Bastani and Glen Berseth and Jeannette Bohg and Ken Goldberg and Abhinav Gupta and Abhishek Gupta and Dinesh Jayaraman and Joseph J Lim and Jitendra Malik and Roberto Martín-Martín and Subramanian Ramamoorthy and Dorsa Sadigh and Shuran Song and Jiajun Wu and Michael C. Yip and Yuke Zhu and Thomas Kollar and Sergey Levine and Chelsea Finn},
      year={2025},
      eprint={2403.12945},
      archivePrefix={arXiv},
      primaryClass={cs.RO},
      url={https://arxiv.org/abs/2403.12945}, 
}

@InProceedings{pi05,
  title = 	 {$\pi_{0.5}$: a Vision-Language-Action Model with Open-World Generalization},
  author =       {Black, Kevin and Brown, Noah and Darpinian, James and Dhabalia, Karan and Driess, Danny and Esmail, Adnan and Equi, Michael Robert and Finn, Chelsea and Fusai, Niccolo and Galliker, Manuel Y. and Ghosh, Dibya and Groom, Lachy and Hausman, Karol and Ichter, Brian and Jakubczak, Szymon and Jones, Tim and Ke, Liyiming and LeBlanc, Devin and Levine, Sergey and Li-Bell, Adrian and Mothukuri, Mohith and Nair, Suraj and Pertsch, Karl and Ren, Allen Z. and Shi, Lucy Xiaoyang and Smith, Laura and Springenberg, Jost Tobias and Stachowicz, Kyle and Tanner, James and Vuong, Quan and Walke, Homer and Walling, Anna and Wang, Haohuan and Yu, Lili and Zhilinsky, Ury},
  booktitle = 	 {Proceedings of The 9th Conference on Robot Learning},
  pages = 	 {17--40},
  year = 	 {2025},
  editor = 	 {Lim, Joseph and Song, Shuran and Park, Hae-Won},
  volume = 	 {305},
  series = 	 {Proceedings of Machine Learning Research},
  month = 	 {27--30 Sep},
  publisher =    {PMLR},
  url = 	 {https://proceedings.mlr.press/v305/black25a.html}
}

@article{zhu2022viola,
  title={VIOLA: Imitation Learning for Vision-Based Manipulation with Object Proposal Priors},
  author={Zhu, Yifeng and Joshi, Abhishek and Stone, Peter and Zhu, Yuke},
  journal={arXiv preprint arXiv:2210.11339},
  doi={10.48550/arXiv.2210.11339},
  year={2022}
}

@article{JMLR:v9:vandermaaten08a,
  author  = {Laurens van der Maaten and Geoffrey Hinton},
  title   = {Visualizing Data using t-SNE},
  journal = {Journal of Machine Learning Research},
  year    = {2008},
  volume  = {9},
  number  = {86},
  pages   = {2579--2605},
  url     = {http://jmlr.org/papers/v9/vandermaaten08a.html}
}

\clearpage
\appendix

\section{Implementation Details for VLM Mid-training}
\label{app:impl_detail_vlm}

We apply a simple-yet-effective learnable scoring function $f(\cdot)$ described in Section~\ref{sec:midtrain} as a single linear layer on top of the frozen VLM's last-token last-hidden-state representation, trained with binary cross-entropy loss. For proximity estimator training, we apply a batch size of 128.
The training data are sampled in a balanced manner from the VLM candidate pool (negative sample) and the target VLA data (positive sample), as described in Section~\ref{sec:setup}. Validation accuracy is computed on a held-out split of the union of VLM negatives and VLA positives.
We apply early-stopping at ${\ge}90\%$ validation accuracy, and experiments show that our binary classifier training usually stops at 75 to 100 steps. We then run data selection process with this proximity estimator on all VLM data to select a subset of top 1280k samples by proximity score for further VLM training.

In VLM mid-training, we leverage the LLaMA-Factory~\citep{zheng2024llamafactory} as the training framework. We perform full-parameter supervised fine-tuning on InternVL3.5-1B, unfreezing the vision encoder, multi-modal projector, and language model. Table~\ref{tab:midtrain_hparams} summarizes the key hyperparameters.

\begin{table}[htbp]
\centering
\caption{Key hyperparameters for VLM mid-training.}
\label{tab:midtrain_hparams}
\small
\begin{tabular}{l c}
\toprule
\textbf{Configuration} & \textbf{Mid-training} \\
\midrule
VLM backbone & InternVL3.5-1B \\
Trainable parameters & All (full fine-tuning) \\
Sequence length & 1024 \\
\midrule
Optimizer & AdamW \\
Peak learning rate & $5 \times 10^{-5}$ \\
Learning rate schedule & Cosine decay \\
Warmup ratio & 0.1 \\
Numerical precision & bfloat16 \\
\midrule
Training steps & 5{,}000 \\
Global batch size & 256 \\
\bottomrule
\end{tabular}
\end{table}

\section{Implementation Details for Alternative Proximity Measurements}
\label{app:impl_proximity}

We provide formal definitions and implementation details for the three alternative proximity measurements compared against our learnable proximity estimator in Section~\ref{sec:ablations}.

\noindent\textbf{Feature-space Average Distance.}
For each candidate VLM sample $x$, we compute its average $\ell_2$ distance to all VLA samples in the frozen VLM representation space:
\begin{equation}
    d_{\mathrm{avg}}(x) = \frac{1}{|\mathcal{D}_{\mathrm{VLA}}|} \sum_{y \in \mathcal{D}_{\mathrm{VLA}}} \left\| \phi(x) - \phi(y) \right\|_2,
\end{equation}
where $\phi(\cdot)$ denotes the last hidden state of the frozen VLM, following the same feature space used in Section~\ref{sec:gap}. Samples with smaller $d_{\mathrm{avg}}(x)$ are ranked as closer to the VLA distribution and selected first.

\noindent\textbf{VLA-conditioned Perplexity.}
We fine-tune the VLM on converted in-domain VLA data in which robot actions are represented as text tokens, yielding a VLA-conditioned model with parameters $\theta_{\mathrm{VLA}}$. The perplexity of a candidate sample $x = (x_1, \ldots, x_T)$ under this model is:
\begin{equation}
    \mathrm{PPL}_{\mathrm{VLA}}(x) = \exp\!\left(-\frac{1}{T}\sum_{t=1}^{T} \log p_{\theta_{\mathrm{VLA}}}(x_t \mid x_{<t})\right).
\end{equation}
Samples with lower $\mathrm{PPL}_{\mathrm{VLA}}(x)$ are considered more compatible with the VLA domain and selected first.

\noindent\textbf{Delta Perplexity.}
Delta perplexity measures the perplexity change induced by VLA fine-tuning:
\begin{equation}
    \Delta\mathrm{PPL}(x) = \mathrm{PPL}_{\mathrm{VLA}}(x) - \mathrm{PPL}_{\mathrm{VLM}}(x),
\end{equation}
where $\mathrm{PPL}_{\mathrm{VLM}}(x)$ is the perplexity under the original pretrained VLM. A more negative $\Delta\mathrm{PPL}(x)$ indicates that the sample becomes more predictable after VLA adaptation, suggesting stronger alignment with the VLA domain. Samples are ranked by $\Delta\mathrm{PPL}(x)$ in ascending order for selection.

\section{Implementation Details for VLA Fine-tuning}
\label{app:impl_detail_vla}

\noindent\textbf{Training Setup.}
We follow the VLM4VLA adaptation design and evaluation protocol~\citep{zhang2026vlm4vla} for downstream VLA adaptation. Table~\ref{tab:vla_setup} summarizes the key training settings for our InternVL3.5-1B-based VLA models. Across all experiments, models take a single-view image as visual input and do not use robot state information, and are fine-tuned end-to-end. Input images are first standardized to $224\times224$ and then resized to match the input resolution required by each VLM backbone. For InternVL3.5-1B, the final input resolution is $448 \times 448$.

For the baseline models reported in Table~\ref{tab:main_results} including expert VLAs and off-the-shelf VLMs, we directly use the results from VLM4VLA.

\begin{table}[htbp]
\centering
\caption{Key hyperparameters for VLA fine-tuning.}
\label{tab:vla_setup}
\small
\begin{tabular}{l c}
\toprule
\textbf{Configuration} & \textbf{VLA Fine-tuning} \\
\midrule
Trainable parameters & All (full fine-tuning) \\
\midrule
\makecell[l]{Action chunk size} & \makecell[c]{10 (Calvin ABC-D) \\ 4 (Simpler-Bridge, Libero-10)} \\
\midrule
Optimizer & AdamW \\
Peak learning rate & $5 \times 10^{-5}$ \\
Learning rate schedule & Constant after linear warmup \\
Warmup steps & 500 \\
Weight decay & 0 \\
Gradient clipping & 1.0 \\
Numerical precision & bfloat16 \\
\midrule
\makecell[l]{Training steps} & \makecell[c]{8,000 (Calvin ABC-D) \\ 16,000 (Simpler-Bridge, Libero-10)} \\
\makecell[l]{Global batch size} & \makecell[c]{128 (Calvin ABC-D) \\ 256 (Simpler-Bridge, Libero-10)} \\
\bottomrule
\end{tabular}

\end{table}

\noindent\textbf{VLA Architecture.}
Following VLM4VLA~\citep{zhang2026vlm4vla}, we append a single learnable action-query embedding $\mathbf{q}$ to the image--instruction sequence. At timestep $t$, the current image $I_t$ is encoded into visual tokens $\mathbf{V}_t$ by the VLM's vision encoder and multimodal projector. These tokens and the task instruction $\ell$ are arranged using the backbone-specific template $\mathcal{T}_m$. The final hidden state at the action-query position provides the control representation:
\begin{equation}
    \mathbf{x}_t = [\mathcal{T}_m(\mathbf{V}_t,\ell);\mathbf{q}],
    \qquad
    \mathbf{h}_t = F_{\theta}(\mathbf{x}_t)_{\mathbf{q}} \in \mathbb{R}^{d},
\end{equation}
where $F_{\theta}$ is the VLM's language backbone and $d$ is its hidden size. For InternVL3.5 and Qwen3VL, the image and instruction are placed in the user message, and the action query follows the assistant prefix.

A lightweight two-branch MLP decodes $\mathbf{h}_t$ into continuous arm actions and binary gripper states for each action chunk. We train the backbone, query embedding, and action head jointly, using Huber loss for normalized arm actions and binary cross-entropy on the gripper logits.

\raggedbottom % avoid stretched vertical gaps before the non-floating Figure in Appendix F
\section{Implementation Details for Real World Experiments}
\label{app:impl_detail_real_world}

\textbf{Hardware Setup.}  We collect demonstrations using a 3D Connexion SpaceMouse
for teleoperation similar to the DROID~\citep{khazatsky2025droidlargescaleinthewildrobot} set up, consisting of
a 7-DoF Franka Research 3 robot with a Robotiq 2F-85 gripper, and an Intel Realsense D435i
external camera. We collect 50 demonstrations for each of five tabletop manipulation
tasks and evaluate with 10 rollouts per task.

\textbf{Training Configuration.} We adapt VLM4VLA's codebase on our real-world dataset, and the training configuration is provided in Table~\ref{tab:franka_vla_setup}.

\begin{table}[htbp]
\centering
\caption{Key hyperparameters for VLA fine-tuning on the Franka real-robot data.}
\label{tab:franka_vla_setup}
\small
\begin{tabular}{l c}
\toprule
\textbf{Configuration} & \textbf{Real-world Fine-tuning} \\
\midrule
Trainable parameters & All (full fine-tuning) \\
\midrule
Action chunk size & 4 \\
Action dimension & 7 (3 transl. + 3 rot.-vec. + gripper) \\
\midrule
Optimizer & AdamW \\
Peak learning rate & $5 \times 10^{-5}$ \\
Learning rate schedule & Constant after linear warmup \\
Warmup steps & 500 \\
Weight decay & 0 \\
Gradient clipping & 1.0 \\
Numerical precision & bfloat16 \\
\midrule
Training steps & 2{,}000  \\
Global batch size & 256 \\
\bottomrule
\end{tabular}
\end{table}

\textbf{Deployment Setting.} We use \textsc{Deoxys Control}~\citep{zhu2022viola} as the real-time control interface for executing VLA-predicted actions on the Franka robot arm. Individual actions from the predicted chunks are supplied at 20~Hz, while \textsc{Deoxys Control} performs low-level control within the robot's 1~kHz control loop.

\medskip
\noindent\begin{minipage}{\linewidth}
\captionsetup{type=table}
\centering
\caption{Text instructions used in the real-world evaluations. We collect 50 demonstrations per task and evaluate each task over 10 rollouts, with a limit of 300 steps per rollout. Representative executions are shown in Figures~\ref{fig:realworld_rollout} and~\ref{fig:realworld_shared_outcomes}.}
\label{tab:realworld_task_definitions}
\small
\renewcommand{\arraystretch}{1.15}
\setlength{\tabcolsep}{10pt}
\begin{tabular}{ll}
\toprule
\textbf{Task} & \textbf{Text Description} \\
\midrule
Cube to plate & put the red cube on the plate \\
\addlinespace
Carrot to plate & put the carrot on the plate \\
\addlinespace
Burger to plate & put the hamburger on the plate \\
\addlinespace
Cube to left cup & put the red cube in the cup on the left \\
\addlinespace
Stack cubes & stack the red cube on top of the green cube \\
\bottomrule
\end{tabular}
\end{minipage}\par
\medskip

\section{Detailed Experimental Results}
\label{app:detailed_results}

Table~\ref{tab:app_simpler_detailed} provides full per-scene breakdowns for SimplerEnv-Bridge, corresponding to the aggregate scores reported in Table~\ref{tab:main_results}. EmbodiedMidtrain improves most of the subtasks consistently.

\begin{table}[htbp]
\centering
\caption{Detailed results on SimplerEnv-Bridge (per-scene success rates: Carrot, Eggplant, Spoon, Cube). * Results for Expert VLA Baselines and Off-the-shelf VLM Baselines are reproduced and reported by VLM4VLA.}
\label{tab:app_simpler_detailed}
\fontsize{8}{9}\selectfont
\setlength{\tabcolsep}{4pt}
\begin{tabular*}{\linewidth}{@{\extracolsep{\fill}}l r r r r r r@{}}
\toprule
\textbf{Model} & \textbf{Size} & \textbf{Carrot} & \textbf{Eggplant} & \textbf{Spoon} & \textbf{Cube} & \textbf{Simpler$\uparrow$} \\
\midrule
\rowcolor{gray!20}\multicolumn{7}{c}{\textbf{Expert VLA Baselines*}}\\
\midrule
OpenVLA (Llama-2) & 7.7B & 4.2 & 0.0 & 0.0 & 12.5 & 4.2 \\
$\pi_0$ (Paligemma-1) & 3.1B & 62.5 & 100.0 & 54.2 & 25.0 & 60.4 \\
\midrule
\rowcolor{gray!20}\multicolumn{7}{c}{\textbf{Off-the-shelf VLM Baselines*}}\\
\midrule
Qwen2.5VL-3B & 3.8B & 20.8 & 91.7 & 79.2 & 0.0 & 48.0 \\
Qwen2.5VL-7B & 8.3B & 12.5 & 100.0 & 75.0 & 0.0 & 46.8 \\
\midrule
Qwen3VL-4B & 4.4B & 54.2 & 95.8 & 75.0 & 0.0 & 56.3 \\
Qwen3VL-8B & 8.8B & 58.3 & 95.8 & 79.2 & 0.0 & 58.3 \\
Qwen3VL-30B-A3B & 30B-A3B & 29.2 & 79.2 & 70.8 & 0.0 & 44.8 \\
\midrule
Paligemma-1 & 2.9B & 50.0 & 91.7 & 75.0 & 4.2 & 55.3 \\
Paligemma-2 & 3.0B & 75.0 & 75.0 & 79.2 & 0.0 & 57.3 \\
KosMos-2 & 1.7B & 37.5 & 100.0 & 75.0 & 29.2 & 60.4 \\
\midrule
\rowcolor{gray!20}\multicolumn{7}{c}{\textbf{VLM with \textit{EmbodiedMidtrain} (Ours)}}\\
\midrule
Qwen3VL-2B & 2.1B & 37.5 & 75.0 & 41.7 & 0.0 & 38.5 \\
\quad \textit{+EmbodiedMidtrain} & 2.1B & 41.7 & 95.8 & 45.8 & 0.0 & 45.8 \\
InternVL3.5-1B & 1.1B & 0.0 & 100.0 & 45.8 & 0.0 & 36.5 \\
\quad \textit{+EmbodiedMidtrain} & 1.1B & 20.8 & 91.7 & 70.8 & 41.7 & \textbf{56.3} \\
\bottomrule
\end{tabular*}

\end{table}

\section{Selected Data Mixture Composition}
\label{app:selected_mixture}

\noindent\begin{minipage}{\linewidth}
    \captionsetup{type=figure}
    \centering
    \includegraphics[width=0.85\linewidth]{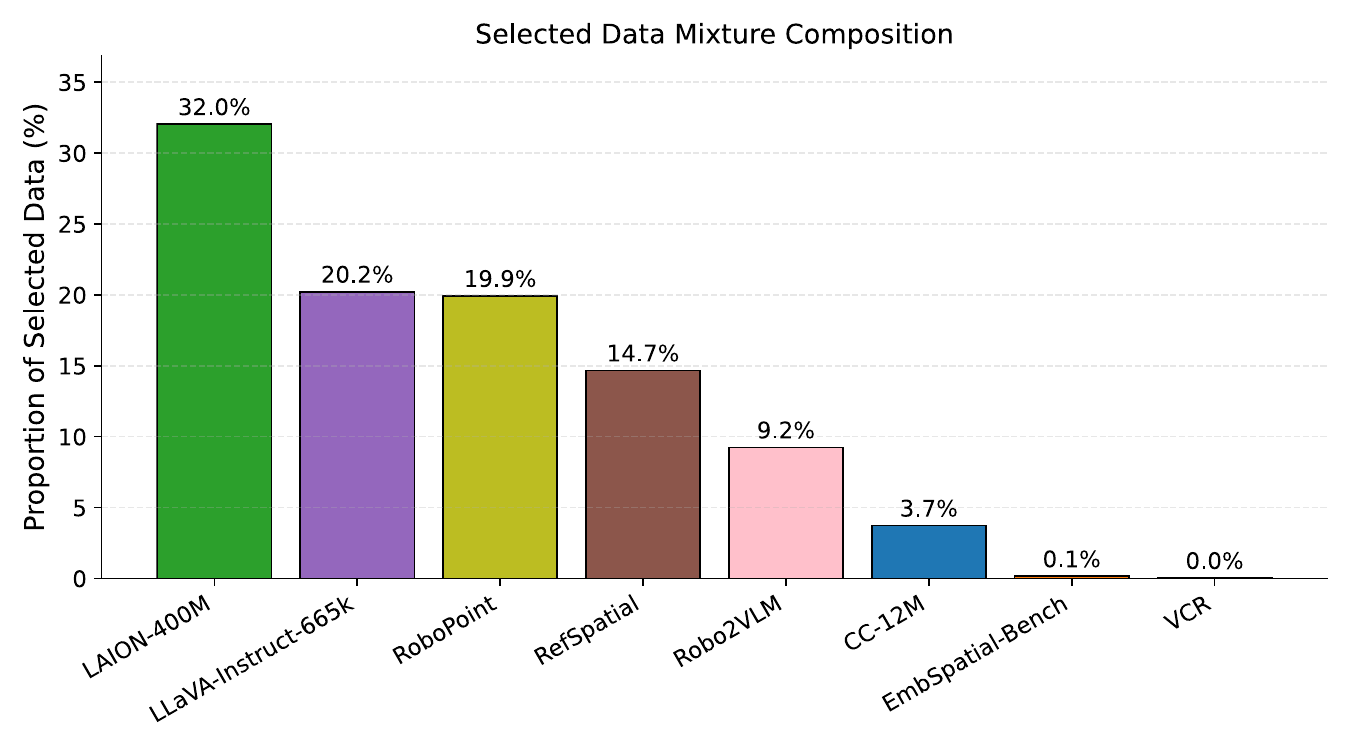}
    \caption{Composition of the selected mid-training data mixture after proximity-based selection, shown as the percentage of each VLM source dataset in the final mixture.}
    \label{fig:selected_mixture}
\end{minipage}\par

Figure~\ref{fig:selected_mixture} shows the composition of the mid-training mixture after proximity-based data selection. LAION-400M contributes the largest share, while EmbSpatial-Bench and VCR are nearly absent from the selected mixture.

\flushbottom
\subsection{Diversity preservation after data selection}
\label{sec:diversity}

We further verify that our proximity-based selection preserves the diversity of the original VLM dataset rather than collapsing the mid-training corpus onto a narrow region near the VLA domain. To quantify the diversity of a data subset, we follow~\citet{xing2025shortcut} to adopt the \emph{uniformity} metric proposed by~\citet{diversity}. Formally, given a dataset $D_i$, we define its diversity as
\begin{equation}
    S_{\text{diversity}}^{D_i} \triangleq \frac{1}{\mathbb{E}_{u, v \sim D_i} \left[ e^{-t \| u - v \|_2^2} \right]},
\end{equation}
where $u$ and $v$ are feature representations drawn from $D_i$ in the frozen VLM's last hidden state space (following Section~\ref{sec:gap}), and $t$ is a temperature parameter controlling the kernel bandwidth (we apply $t=2$ in our calculation). A higher $S_{\text{diversity}}^{D_i}$ indicates greater spread among samples, reflecting higher diversity.

\begin{table}[t]
    \centering
    \caption{Diversity of VLM data, VLA data, and the VLM subset selected by EmbodiedMidtrain. The full VLM data pool comprises two disjoint subsets: general VLM data and embodied-oriented VLM data.}
    \label{tab:diversity}
    \small
    \setlength{\tabcolsep}{6pt}
    \begin{tabular}{@{}lr@{}}
        \toprule
        \textbf{Dataset} & \textbf{Diversity} \\
        \midrule
        VLM data & 1.85 \\
        \quad \textit{General VLM Data} & 1.96 \\
        \quad \textit{Embodied-oriented VLM Data} & 1.62 \\
        \addlinespace
        VLA data & 1.26 \\
        \midrule
        Selected VLM data & 1.93 \\
        \bottomrule
    \end{tabular}
\end{table}

Table~\ref{tab:diversity} reports the diversity scores. VLA data is the most concentrated, with a score of 1.26 that is consistent with the compact clusters observed in Section~\ref{sec:gap}, while the general VLM pool is the most diverse at 1.96. Our selected subset reaches 1.93, closely matching the general VLM pool and substantially exceeding both the embodied-oriented subset at 1.62 and the VLA data. Proximity-based selection thus shifts the data distribution toward the VLA domain without collapsing it, retaining the broad visual and linguistic coverage that makes the mixture effective as a mid-training signal.

\section{VLM Capability After Mid-training}
\label{app:vlm_retention}

While our primary goal is to improve VLMs as initializations for downstream VLA adaptation, it is also important to examine how mid-training affects the original capability profile of the VLM itself. To this end, we evaluate the mid-trained model on VLM benchmarks including BLINK~\citep{fu2024blinkmultimodallargelanguage}, POPE~\citep{POPE}, VisuLogic~\citep{xu2025visulogic}, 3DSRBench~\citep{3DSRBench} and SpatialEval~\citep{spatialeval}. Results are reported in Table~\ref{tab:vlm_retention}.

Overall, mid-training largely preserves the original VLM capability profile while inducing a selective shift in performance. Performance remains nearly unchanged on POPE, improves on VisuLogic and 3DSRBench, and decreases moderately on BLINK and SpatialEval. This suggests that VLM mid-training does not uniformly maintain all original capabilities, but instead reorients the model toward skills that are more relevant to embodied downstream adaptation.

\begin{table}[tbp]
\centering
\caption{Performance of InternVL3.5-1B on VLM benchmarks before and after mid-training.}
\label{tab:vlm_retention}
\small
\setlength{\tabcolsep}{2.5pt}
\begin{adjustbox}{max width=0.7\linewidth}
\begin{tabular}{lccccc}
\toprule
\textbf{Model} & \textbf{POPE$\uparrow$} & \textbf{BLINK$\uparrow$} & \textbf{VisuLogic$\uparrow$} & \textbf{3DSRBench$\uparrow$} & \textbf{SpatialEval$\uparrow$} \\
\midrule
InternVL3.5-1B          & 86.33 & 43.45 & 21.00 & 47.87 & 49.82 \\
\quad \textit{+EmbodiedMidtrain}       & 86.29 & 40.45 & 24.90 & 49.51 & 48.00 \\
\bottomrule
\end{tabular}
\end{adjustbox}

\end{table}

\noindent\textbf{Detailed BLINK subtask results.} We examine performance across BLINK subtasks to understand which capabilities change after EmbodiedMidtrain. Table~\ref{tab:blink_subtask} reports the five subtasks with the largest changes: two with declines and three with gains. The declines occur in tasks that require reasoning across multiple images or jointly presented image patches. These input formats are absent from our robotics fine-tuning data and rare in the original VLM pool, leaving few such examples in the selected mid-training data. This limited coverage may contribute to the observed declines. The three subtasks that improve (Spatial Relation, Counting, and Semantic Correspondence) involve spatial and object-level reasoning, consistent with the emphasis of our data selection. The results suggest a trade-off associated with the composition of the mid-training data, with the largest losses concentrated in input formats that are uncommon in robotics. Our objective is to improve VLA initialization, rather than to improve performance uniformly across all general-purpose VLM tasks.

\begin{table}[htbp]
\centering
\caption{Top-5 most affected BLINK sub-tasks before and after mid-training. The two largest drops occur on multi-image / image-mixing tasks, and the gains occur on spatial- and object-grounded tasks.}
\label{tab:blink_subtask}
\small
\begin{tabular}{l c c c}
\toprule
\textbf{BLINK sub-task} & \textbf{Pre} & \textbf{Post} & \textbf{$\Delta$} \\
\midrule
Visual Similarity         & 0.726 & 0.474 & $-0.252$ \\
Jigsaw                    & 0.620 & 0.527 & $-0.093$ \\
Spatial Relation          & 0.678 & 0.706 & $+0.028$ \\
Counting                  & 0.492 & 0.583 & $+0.091$ \\
Semantic Correspondence   & 0.230 & 0.345 & $+0.115$ \\
\bottomrule
\end{tabular}

\end{table}

\section{Transferability across Target Tasks and VLA Action Heads}
\label{app:transferability}

\subsection{Transferability across target tasks}
\label{app:ood_selector}

By construction, \textbf{no robot manipulation data used for downstream VLA fine-tuning are included in the VLM mid-training dataset}: the proximity classifier uses only a small subset of data from three simulation VLA datasets as target data for selection. Real-robot results in Section~\ref{sec:real_world_exp} show that our mid-training gains transfer from simulation to real-world tasks: the dataset selected using simulation data as the target consistently benefits downstream fine-tuning in the real world. In addition, we conduct leave-one-benchmark-out experiments in this section to provide further evidence of transferability across target tasks.

\noindent\textbf{Leave-one-benchmark-out generalization.}
We train the proximity classifier using \emph{only} the Bridge and Calvin VLA training data, holding LIBERO-10 entirely out of the selector. We then re-run sample selection, mid-train an InternVL3.5-1B with the selected mixture, and VLA-fine-tune on LIBERO. Table~\ref{tab:app_ood} reports the result: the held-out variant retains 95\% of the full method's gain on LIBERO. This indicates that the selector identifies VLM samples useful for embodied tasks in general, rather than retrieving samples that resemble a specific evaluation benchmark.

\begin{table}[t]
    \centering
    \caption{Leave-one-benchmark-out generalization on LIBERO-10.
    All settings use InternVL3.5-1B. Excluding LIBERO from the
    selection target datasets retains most of the gain from mid-training.}
    \label{tab:app_ood}
    \small
    \setlength{\tabcolsep}{4pt}
    \begin{tabular}{@{}llc@{}}
        \toprule
        \textbf{Method} & \textbf{Selection targets} & \textbf{LIBERO-10} \\
        \midrule
        No mid-training & N/A & 39.0 \\
        \textit{EmbodiedMidtrain} & Bridge + CALVIN & 53.4 \\
        \textit{EmbodiedMidtrain} & Bridge + CALVIN + LIBERO & \textbf{54.2} \\
        \bottomrule
    \end{tabular}
\end{table}

\subsection{Transferability across VLA action heads}
\label{app:action_head}

In the main experiments, we follow VLM4VLA's architecture design using two-branch MLP action head. To demonstrate that the performance gain of EmbodiedMidtrain can transfer across different action head design, we additionally evaluate EmbodiedMidtrain with two qualitatively different action heads using a modular VLA training framework that supports multiple decoder heads: parallel decoding of OpenVLA-OFT~\citep{kim2025oft} style on SimplerEnv-Bridge, and flow matching of $\pi_0$~\citep{black2024pi0} style on LIBERO-10. We evaluate both InternVL3.5-1B and Qwen3VL-2B on this transferability. Experiment result is shown in Table~\ref{tab:app_action_head}.

\begin{table}[t]
    \centering
    \caption{Transfer across action heads. EmbodiedMidtrain
    consistently improves both parallel decoding and flow matching
    across two backbones, with gains of 4.0--12.5 points.
    These results suggest that its benefits extend beyond the
    two-branch MLP head used in the main experiments.}
    \label{tab:app_action_head}
    \small
    \setlength{\tabcolsep}{6pt}
    \begin{adjustbox}{max width=\linewidth}
    \begin{tabular}{@{}lcc@{}}
        \toprule
        \textbf{Setting} & \textbf{Parallel decoding} & \textbf{Flow matching} \\
        & SimplerEnv & LIBERO-10 \\
        \midrule
        InternVL3.5-1B & 36.5 & 88.4 \\
        \quad + \textit{EmbodiedMidtrain} & \textbf{49.0} & \textbf{92.4} \\
        \midrule
        Qwen3VL-2B & 42.7 & 83.4 \\
        \quad + \textit{EmbodiedMidtrain} & \textbf{50.0} & \textbf{87.6} \\
        \bottomrule
    \end{tabular}
    \end{adjustbox}
\end{table}

EmbodiedMidtrain consistently improves performance across VLA action heads for both VLM backbones. Together with the main results on VLM4VLA's two-branch MLP architecture in Table~\ref{tab:main_results}, these findings demonstrate that the performance gains from mid-training transfer across different VLA action heads.

\section{Compute Accounting and FLOP-Matched Comparison}
\label{app:compute_flop}

\noindent\textbf{Compute accounting.}
The mid-training stage costs approximately \textbf{60 GPU-hours} on NVIDIA L40S (5{,}000 steps at global batch size 256, full-parameter optimization). Each VLA fine-tuning run on the same hardware costs approximately \textbf{300 GPU-hours}. Mid-training therefore adds about 20\% to a single VLA fine-tuning run. However, the mid-trained VLM checkpoint is a one-time cost that is amortized across multiple downstream VLA runs (different VLA datasets, different action heads, etc.). Across the experiments in this paper, the same mid-trained checkpoint is reused for more than ten downstream VLA runs, so the effective per-run overhead is much smaller than the 20\% headline number suggests.

\noindent\textbf{FLOP-matched comparison.}
To rule out the possibility that the gain is a compute artifact, we train an off-the-shelf InternVL3.5-1B baseline for additional VLA fine-tuning steps so that its total compute matches mid-training plus standard VLA fine-tuning. Table~\ref{tab:app_flop} reports the comparison. At identical total FLOPs, EmbodiedMidtrain substantially outperforms simply spending the additional budget on extended VLA fine-tuning ($+12.5$ on SimplerEnv, $+8.4$ on LIBERO-10). The gain is therefore not a compute artifact: at matched total compute, mid-training is the more effective use of additional budget.

\begin{table}[htbp]
\centering
\caption{FLOP-matched comparison. The baseline is trained for extra VLA fine-tuning steps so that its total compute equals mid-training plus VLA fine-tuning. At identical total FLOPs, EmbodiedMidtrain substantially outperforms extended VLA fine-tuning.}
\label{tab:app_flop}
\small
\setlength{\tabcolsep}{4pt}
\begin{adjustbox}{max width=\linewidth}
\begin{tabular}{@{}>{\raggedright\arraybackslash}p{0.65\linewidth} c c@{}}
\toprule
\textbf{Setting} & \textbf{SimplerEnv} & \textbf{LIBERO-10} \\
\midrule
InternVL3.5-1B (extra training to match total compute) & 43.8 & 45.8 \\
InternVL3.5-1B + \textit{EmbodiedMidtrain} & \textbf{56.3} & \textbf{54.2} \\
\bottomrule
\end{tabular}
\end{adjustbox}

\end{table}

\section{Proximity Selector Design Ablations}
\label{app:selector_ablations}

\begin{table}[t]
    \centering
    \caption{Effect of selection size $K$ on downstream VLA performance with InternVL3.5-1B. All mid-training variants use the same compute budget. Baseline denotes the model without mid-training. Full pool uses all candidate data without selection.}
    \label{tab:app_k_sweep}
    \small
    \setlength{\tabcolsep}{4pt}
    \begin{adjustbox}{max width=\linewidth}
    \begin{tabular}{@{}lccccc@{}}
        \toprule
        \textbf{Benchmark} & \textbf{Baseline}
        & \multicolumn{3}{c}{\textbf{EmbodiedMidtrain} ($K$)}
        & \textbf{Full pool} \\
        \cmidrule(lr){3-5}
        & & $320\,\mathrm{k}$ & $1.28\,\mathrm{M}$
          & $2.56\,\mathrm{M}$ & \\
        \midrule
        CALVIN
            & 3.173 & 3.211 & \textbf{3.714} & 3.455 & 3.398 \\
        SimplerEnv
            & 36.5 & 47.9 & \textbf{56.3} & 51.0 & 43.8 \\
        LIBERO-10
            & 39.0 & 48.6 & \textbf{54.2} & 50.4 & 48.4 \\
        \bottomrule
    \end{tabular}
    \end{adjustbox}
\end{table}

\noindent\textbf{Selection size.}
We compare selection sizes $K \in \{320\,\mathrm{k}, 1.28\,\mathrm{M}, 2.56\,\mathrm{M}\}$ with mid-training on the full candidate pool, holding mid-training compute fixed. Table~\ref{tab:app_k_sweep} shows gains over the off-the-shelf InternVL3.5-1B initialization for every selected subset. The default selection of $1.28\,\mathrm{M}$ examples yields the best performance on all three benchmarks, followed by $2.56\,\mathrm{M}$, as both of them outperform full-pool mid-training. The $320\,\mathrm{k}$ subset also outperforms the full pool on SimplerEnv and LIBERO-10, but falls below it on CALVIN.

\noindent\textbf{Classifier stopping point.}
We stop training the proximity classifier when held-out classification accuracy reaches at least $90\%$, as described in Appendix~\ref{app:impl_detail_vlm}. To examine how training affects proximity scores, we score the full candidate pool using checkpoints with validation accuracies of approximately $70\%$, $90\%$, and $95\%$. Table~\ref{tab:app_earlystop} reports the within-source interquartile range (IQR) for three VLM sources and the fraction of all candidates with scores below $0.10$. This threshold summarizes the score distribution. Data selection uses top-$K$ ranking throughout.

\begin{table}[htbp]
\centering
\caption{Proximity score statistics at three classifier checkpoints, indexed by validation accuracy. IQR denotes the interquartile range within each source. The final row reports the fraction of the full candidate pool with $s(x)<0.10$.}
\label{tab:app_earlystop}
\small
\begin{tabular}{l c c c}
\toprule
\textbf{Metric} & \textbf{70\%} & \textbf{90\% (ours)} & \textbf{95\%} \\
\midrule
Score IQR, LAION-400M    & 0.293 & 0.158 & 0.042 \\
Score IQR, LLaVA-Instruct    & 0.429 & 0.224 & 0.057 \\
Score IQR, CC-12M   & 0.244 & 0.123 & 0.046 \\
$P(s(x) < 0.10)$             & 21.0\% & 52.8\% & 80.7\% \\
\bottomrule
\end{tabular}

\end{table}

The $70\%$ checkpoint has the largest within-source IQRs and the smallest fraction of low-scoring candidates. Further training concentrates scores near zero: at $95\%$ validation accuracy, $80.7\%$ of candidates have scores below $0.10$, and the IQR is smallest for all three sources. The default $90\%$ checkpoint retains greater within-source score variation than the later checkpoint while satisfying the stopping criterion.

\section{Selection Granularity and Data Source Contributions}
\label{app:selection_decomposition}

We evaluate the effects of selection granularity and candidate data sources using InternVL3.5-1B. Table~\ref{tab:app_sample_vs_dataset} compares dataset-level selection, sample-level selection, and selection restricted to either general or embodied-oriented VLM sources. All mid-training variants use the same compute budget and hyperparameters.

\begin{table}[htbp]
\centering
\caption{Effect of selection granularity and candidate data sources on downstream VLA performance with InternVL3.5-1B.}
\label{tab:app_sample_vs_dataset}
\small
\setlength{\tabcolsep}{4pt}
\begin{adjustbox}{max width=\linewidth}
\begin{tabular}{@{}>{\raggedright\arraybackslash}p{0.64\linewidth} c c c@{}}
\toprule
\textbf{Setting} & \textbf{Calvin} & \textbf{SimplerEnv} & \textbf{LIBERO-10} \\
\midrule
InternVL3.5-1B (no mid-training)                                                 & 3.173 & 36.5 & 39.0 \\
\quad + Mid-training on the full pool (random selection)                                          & 3.398 & 43.8 & 48.4 \\
\quad + Dataset-level selection (no sample filtering)              & 3.273 & 43.8 & 46.6 \\
\quad + Sample-level selection within general VLM sources                            & 3.636 & 49.0 & 51.8 \\
\quad + Sample-level selection within embodied-oriented VLM sources                           & 3.324 & 40.6 & 50.6 \\
\quad + \textit{EmbodiedMidtrain} (sample-level selection across all sources)                             & \textbf{3.714} & \textbf{56.3} & \textbf{54.2} \\
\bottomrule
\end{tabular}
\end{adjustbox}

\end{table}

\noindent\textbf{Dataset-level and sample-level selection.}
Dataset-level selection retains sources with high mean proximity scores without filtering individual samples and yields performance comparable to mid-training on the full pool. Sample-level selection improves over dataset-level selection by $0.441$ in average task sequence length on Calvin and by $12.5$ and $7.6$ percentage points on SimplerEnv and LIBERO-10, respectively. These results support using individual proximity scores to identify useful samples within sources, beyond the information captured by dataset averages.

\noindent\textbf{Contributions of general and embodied-oriented sources.}
Sample-level selection from general VLM sources (LAION-400M, CC-12M, LLaVA-Instruct, and VCR) outperforms selection from embodied-oriented sources (RefSpatial, RoboPoint, Robo2VLM, and EmbSpatial-Bench) on all three benchmarks. Selecting across both groups yields the best results, supporting the use of general and embodied-oriented data together for mid-training.

\section{Qualitative Examples}
\label{app:qualitative}

\subsection{Simulation}

Figure~\ref{fig:qualitative_rollout} shows representative rollouts on the SimplerEnv-Bridge \textit{put carrot on plate} task. For both Qwen3VL-2B and InternVL3.5-1B, the VLA initialized from the off-the-shelf VLM fails to complete the task within the allotted horizon, while the variant initialized from an \textit{EmbodiedMidtrain}-mid-trained VLM successfully grasps the carrot and transports it to the plate. These qualitative outcomes are consistent with the per-scene \textit{Carrot} gains reported in Table~\ref{tab:app_simpler_detailed}, providing visual evidence that proximity-based mid-training yields backbones better suited to embodied control.

\noindent\begin{minipage}{\linewidth}
    \captionsetup{type=figure}
    \centering
    \includegraphics[width=\linewidth]{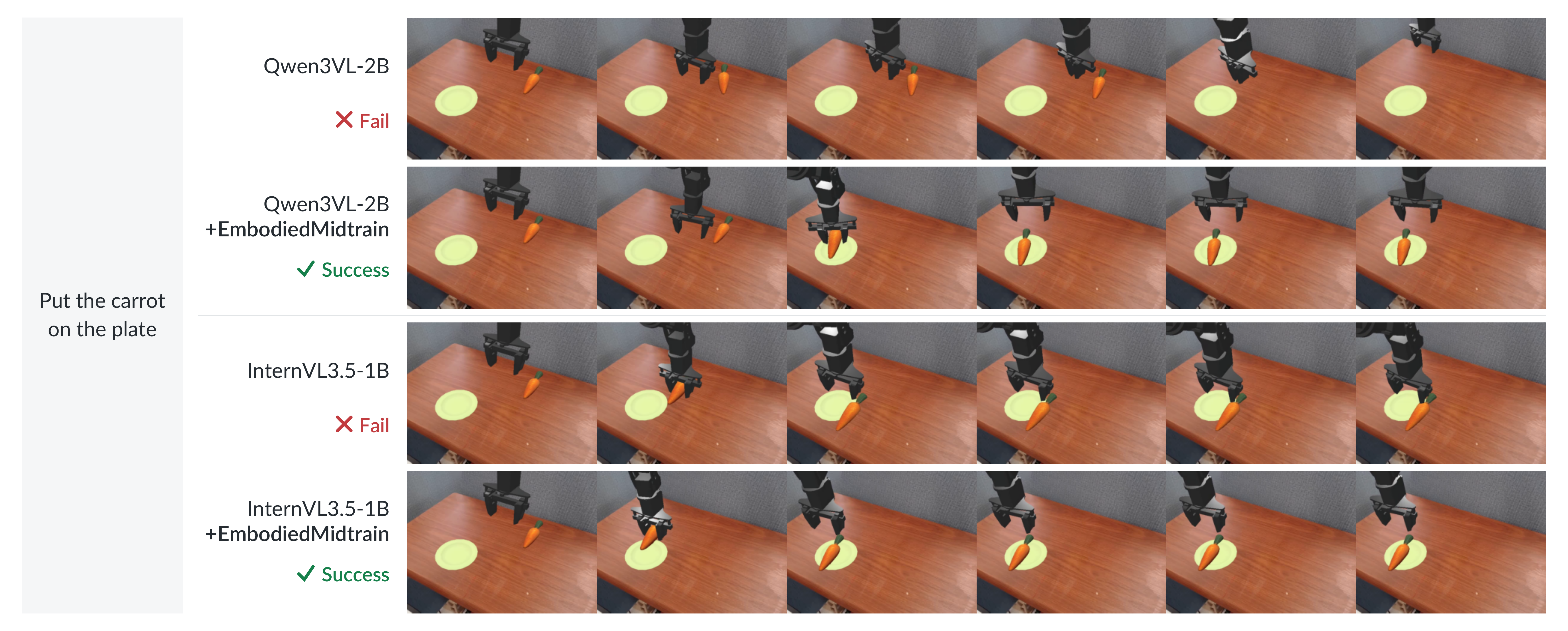}
    \caption{Qualitative rollouts on the SimplerEnv-Bridge \textit{put carrot on plate} task. For each VLM backbone (Qwen3VL-2B and InternVL3.5-1B), we compare the VLA fine-tuned from the off-the-shelf VLM (top of each pair) against the same backbone after \textit{EmbodiedMidtrain} (bottom of each pair). Without mid-training, both baselines fail to complete the task, whereas the mid-trained variants successfully grasp the carrot and place it on the plate.}
    \label{fig:qualitative_rollout}
\end{minipage}\par

\subsection{Real-World Task}
\label{app:qualitative_realworld}

We examine the 50 evaluation rollouts from each initialization across the five tasks in Table~\ref{tab:realworld_task_definitions}. Figure~\ref{fig:realworld_rollout} shows examples where \textit{EmbodiedMidtrain} improves task completion.

\noindent\textbf{Improvements in grasping and placement.}
In the carrot-to-plate example in Figure~\ref{fig:realworld_rollout}, the baseline initialized from InternVL3.5-1B reaches the carrot and closes its gripper too close to the edge, leaving the carrot on the table. With \textit{EmbodiedMidtrain}, the robot successfully lifts the carrot and places it on the plate. Similarly, the baseline displaces the hamburger during its grasp attempt without delivering it to the plate, whereas the mid-trained model maintains its grasp through transport and placement. The cube-to-cup example illustrates a failure after a successful grasp: the baseline carries the cube near the cup but approaches at insufficient height, causing its gripper to push the cup toward the table edge. In the stacking example, the baseline moves the red cube toward the green cube without lifting it high enough to clear the nearby corn and fails to complete the stack. The mid-trained model lifts the red cube above the corn and successfully stacks it on the green cube.

\noindent\textbf{Shared successes.}
Both models succeed in 90\% runs of cube-to-plate task. In Figure~\ref{fig:realworld_shared_outcomes}(a), both policies grasp the red cube, lift it from the table, and release it onto the plate. Shared successes also occur on the carrot and hamburger tasks, where both models complete the same pick-and-place sequence.

\noindent\textbf{Remaining failures.}
In the shared cube-to-plate failure in Figure~\ref{fig:realworld_shared_outcomes}(b), both models approach the cube but leave it on the table without establishing a successful grasp. A possible explanation is the cube's initial position near the edge of the camera view, where visual conditions may differ from those in the regions covered by most collected demonstrations.

\noindent\begin{minipage}{\linewidth}
    \captionsetup{type=figure}
    \centering
    \includegraphics[width=\linewidth]{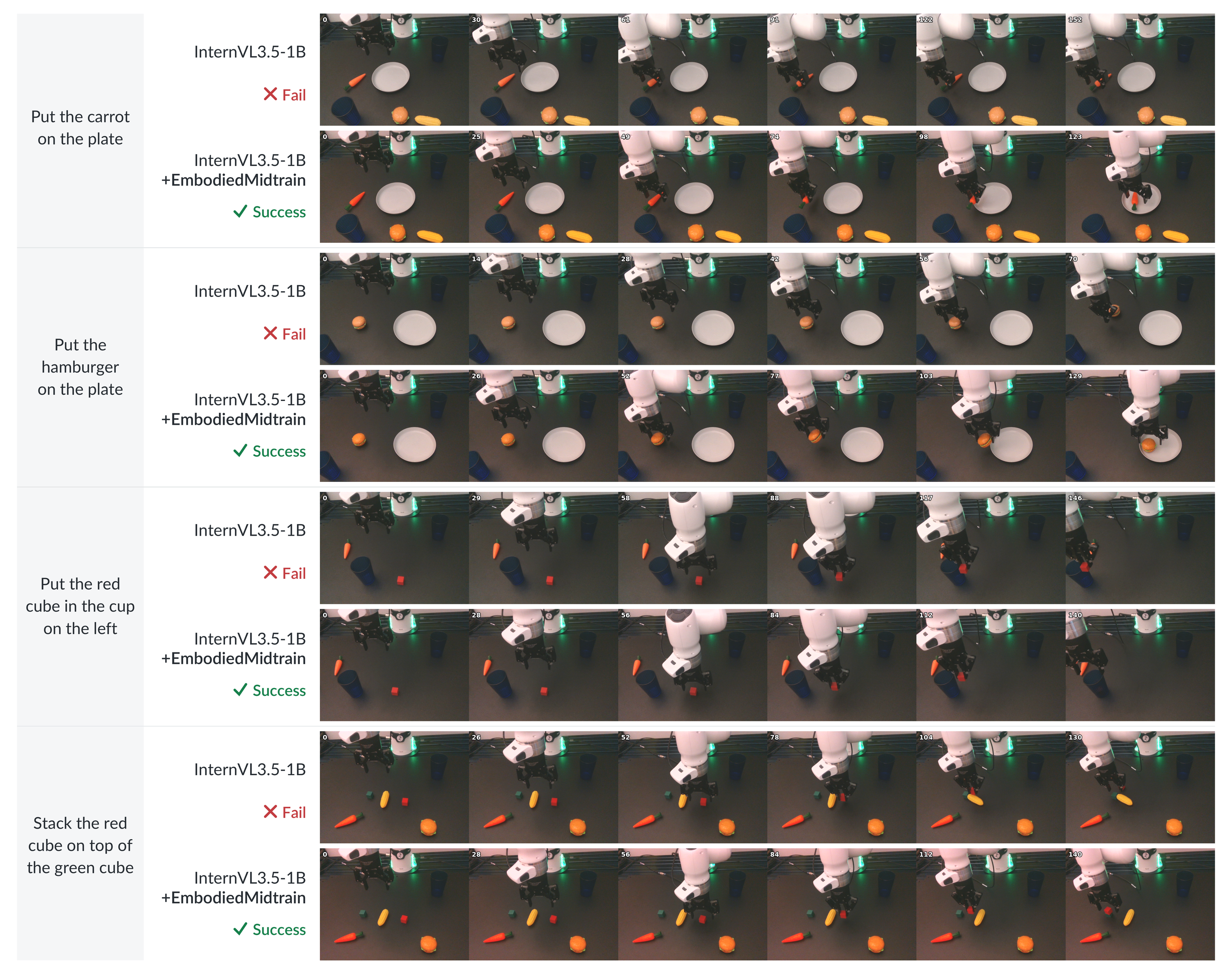}
    \caption{Representative real-world rollouts on InternVL3.5-1B. For each task, the top row shows a failed rollout from the off-the-shelf initialization, and the bottom row shows a successful rollout after \textit{EmbodiedMidtrain}.}
    \label{fig:realworld_rollout}
\end{minipage}\par
\medskip

\medskip

\noindent\begin{minipage}{\linewidth}
\captionsetup{type=figure}
\centering
\includegraphics[width=\linewidth]{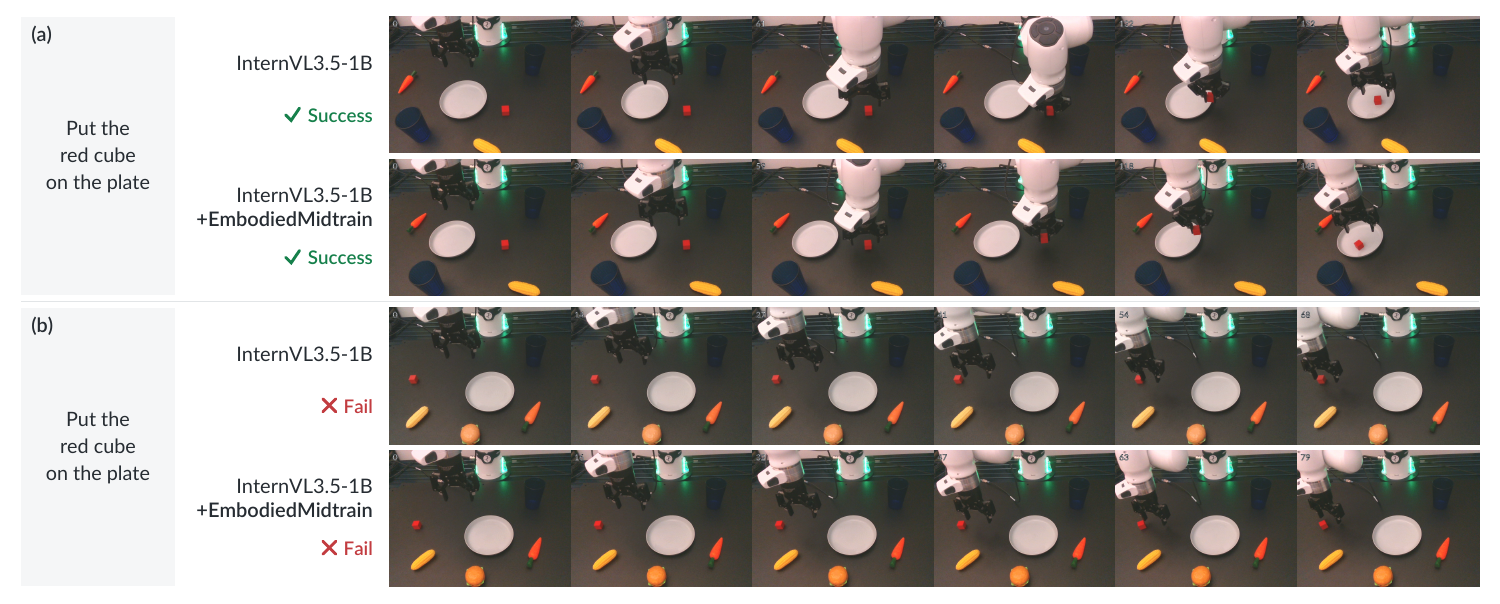}
\caption{Shared successes and failures on the cube-to-plate task. (a) Both models grasp the red cube and place it on the plate. (b) Both approach the cube but leave it on the table. Frames are ordered chronologically within each row.}
\label{fig:realworld_shared_outcomes}
\end{minipage}\par

\end{document}